%% file: sample-manuscript.tex
\documentclass[manuscript,screen]{acmart}
\AtBeginDocument{%
  }

\setcopyright{acmlicensed}
\copyrightyear{2018}
\acmYear{2018}
\acmDOI{XXXXXXX.XXXXXXX}

\acmConference[Conference acronym 'XX]{Make sure to enter the correct
  conference title from your rights confirmation emai}{June 03--05,
  2018}{Woodstock, NY}
\acmISBN{978-1-4503-XXXX-X/18/06}

\usepackage{amssymb}
\usepackage{amsmath}
\usepackage{algorithm} 
\usepackage{algorithmic}
\usepackage{natbib}
\usepackage{bbm}
\usepackage{bm}
\usepackage{setspace}
\usepackage{multirow}
\usepackage{hyperref}

\usepackage{tikz}
\usepackage{graphicx}
\usetikzlibrary{shapes, arrows.meta, positioning}
\usepackage{tabularx}
\usepackage{booktabs}
\usepackage{multirow}
\usepackage{placeins}
\usetikzlibrary{arrows,shapes,positioning}
\usetikzlibrary{calc,decorations.markings}
\usepackage{paralist}

\begin{document}

\title{A Survey of Multimodal Large Language Model from A Data-centric Perspective}

\author{Tianyi Bai}
\authornote{Both authors contributed equally to this research.}
\affiliation{
 \institution{Hong Kong University of Science and Technology}
  \city{Hong Kong}
  \country{China}
}
\email{tbaiag@cse.ust.hk}

\author{Hao Liang}
\authornotemark[1]
\authornote{The work is done during an internship at Apple.}
\affiliation{%
  \institution{Peking University}
  \city{Beijing}
  \country{China}
}
\email{hao.liang@stu.pku.edu.cn}

\author{Binwang Wan}
\affiliation{%
  \institution{Harbin Institute of Technology}
  \city{Weihai}
  \country{China}
}
\email{binwangwan@gmail.com}

\author{Yanran Xu, Xi Li, Shiyu Li}
\affiliation{%
  \institution{Apple}
  \country{China}
}
\email{xu_yanran@apple.com, weston_li@apple.com, shiyu_li@apple.com}

\author{Ling Yang, Bozhou Li}
\affiliation{%
  \institution{Peking University}
  \city{Beijing}
  \country{China}
}
\email{yangling0818@163.com, libozhou@pku.edu.cn}

\author{Yifan Wang}
\affiliation{%
  \institution{University of Science and Technology of China}
  \city{He Fei}
  \country{China}
}
\email{wangyfan@mail.ustc.edu.cn}

\author{Bin Cui}
\affiliation{%
 \institution{Peking University}
 \country{China}
}
\email{bin.cui@pku.edu.cn}

\author{Ping Huang, Jiulong Shan}
\affiliation{%
 \institution{Apple}
 \country{China}
}
\email{huang_ping@apple.com, jlshan@apple.com}

\author{Conghui He}
\authornote{Conghui He, Binhang Yuan and Wentao Zhang are the corresponding authors.}
\affiliation{%
  \institution{Shanghai Artifcial Intelligence Laboratory}
  \country{China}
}
\email{heconghui@pjlab.org.cn}

\author{Binhang Yuan}
\authornotemark[3]
\affiliation{%
 \institution{Hong Kong University of Science and Technology}
 \country{China}
}
\email{biyuan@ust.hk}

\author{Wentao Zhang}
\authornotemark[3]
\affiliation{%
  \institution{Peking University}
  \country{China}
}
\email{wentao.zhang@pku.edu.cn}

\renewcommand{\shortauthors}{Bai et al.}

\begin{abstract}
Multimodal large language models (MLLMs) enhance the capabilities of standard large language models by integrating and processing data from multiple modalities, including text, vision, audio, video, and 3D environments. 
Data plays a pivotal role in the development and refinement of these models. 
In this survey, we comprehensively review the literature on MLLMs from a data-centric perspective.
Specifically, we explore methods for preparing multimodal data during the pretraining and adaptation phases of MLLMs. 
Additionally, we analyze the evaluation methods for the datasets and review the benchmarks for evaluating MLLMs. Our survey also outlines potential future research directions. 
This work aims to provide researchers with a detailed understanding of the data-driven aspects of MLLMs, fostering further exploration and innovation in this field.
\end{abstract}




\maketitle

\input{section/sec1_intro}
\input{section/sec2_overview}

\input{section/sec3_datapro}
\input{section/sec4_pretrain}
\input{section/sec5_adaptation}
\input{section/sec6_evaluation}
\input{section/sec7_future_direction}
\input{section/sec8_conclusion}



\bibliographystyle{ACM-Reference-Format}
\bibliography{reference}

\input{section/appendix1.tex}

\end{document}

%% file: section/sec1_intro.tex
\section{introduction}\label{sec:intro}
In recent years, we have witnessed rapid advancement of large language models (LLMs) and multimodal large language models (MLLMs)~\cite{zhao2023survey,wu2023multimodal}. 
MLLMs, such as GPT-4~\cite{openai2023gpt}, Flamingo~\cite{alayrac2022flamingo}, LLaVA~\cite{liu2024visual}, BLIP2~\cite{li2023blip}, Cambrian-1~\cite{tong2024cambrian} and X-InstructBLIP~\cite{panagopoulou2023x}, integrate multiple modality information, demonstrating impressive comprehension and generation capabilities. 
These models achieve competitive performance in traditional multimodal tasks, such as visual recognition~\cite{zhang2024vision}, video understanding~\cite{xu2021vlm, tang2023video}, speech recognition~\cite{min2023exploring} and 3D understanding~\cite{guo2023point, hong20233d}. Moreover, their excellent language understanding capacity enables strong performance in text-rich tasks, such as question answering~\cite{hu2024bliva}, multi-dialog conversation and logical reasoning~\cite{xu2024llava, li2023m}.

Most existing MLLMs focus on modifying model architecture to explore the use of information from modalities~\cite{xu2024llava, panagopoulou2023x, alayrac2022flamingo}. While model effectiveness is crucial, data also significantly impacts the success of MLLMs.  
For example, \citet{hoffmann2022training} shows that in order to scale up models, it is necessary to increase the scale of the training data. 
Beyond data volume, data quality is equally important. Previous research~\cite{sorscher2022beyond} indicates that carefully curated datasets can enable smaller models to achieve comparable performance to larger ones. However, comprehensive studies on data curation and utilization for MLLMs are still lacking. Therefore, this study aims to provide a comprehensive understanding of MLLMs from a data-centric perspective.

In contrast to model-centric approaches that prioritize architectural enhancements while relying on fixed datasets, data-centric perspectives emphasize the comprehensive impact of the training corpus datasets on the model performance.
Within the scope of data-centric MLLMs, our focus lies on leveraging the heterogeneous nature of data modalities, enhancing data structure, increasing data quantity, and elevating data quality to augment MLLMs~\cite{zha2023data}. Our survey discusses three key questions from a data-centric perspective at different stages of MLLMs:

\begin{itemize}
\item {\em \textbf{Q1:} How to collect, select, and manage data for MLLMs?}  
The substantial data volume requirements and the heterogeneity of multimodal data pose challenges in gathering, selecting, and effectively managing data for model training. Different training stages of MLLMs also lead to varying data type requirements.

\item {\em \textbf{Q2:} How data affects the performance of MLLMs?}
Understanding the relationship between data characteristics and the performance of MLLMs is crucial for optimizing datasets and enhancing model capabilities.

\item {\em \textbf{Q3:} How to evaluate the data for MLLMs?} It is necessary to develop comprehensive evaluation benchmarks to assess the performance and robustness of MLLMs across diverse tasks.
\end{itemize}

\textbf{Unique contribution of this survey.}
Several existing works have focused on LLMs~\cite{zhao2023survey,hadi2023large,naveed2023comprehensive} and MLLMs~\cite{wu2023multimodal,zhang2024mm} from a model-centric perspective, but lack an in-depth analysis from the data-centric perspective. 
Recently, some work has started to focus on data preparation for LLMs, such as data management methods~\cite{wang2023data}, data selection methods~\cite{albalak2024survey}, and comprehensive reviews of LLM datasets~\cite{liu2024datasets}. 
However, such work primarily focuses on data management and selection methods for text-only LLMs and does not provide a detailed analysis of the data processing pipeline for MLLMs. 
The most closely related work to ours is data-centric artificial intelligence (DCAI)~\cite{zha2023data,jarrahi2022principles,jakubik2022data,polyzotis2021can,whang2023data}, which also focuses on data-centric views of AI research but does not specifically analyze LLMs and MLLMs.

With the rapid growth of MLLMs and the increasingly important role of data in this large model era, we believe it is crucial to provide a comprehensive overview of data-centric approaches for MLLMs. This survey aims to thoroughly review the literature on the advances of MLLMs from a data-centric perspective and discusses open issues or future directions in this field.

In this survey, we reviewed the literature on the advances of MLLMs from data-centric perspectives. We provide researchers and developers with a general and comprehensive understanding of the latest developments in the field of MLLM data side. The key contributions of this survey are summarized as follows:
\begin{itemize}
    \item \textit{\textbf{New data-centric perspective.}} We provide a comprehensive review of MLLMs from a data-centric perspective, considering modalities such as text, image, video, and audio.
    
    \item \textit{\textbf{Data preparation and management pipeline.}} We summarize the data preparation and management pipeline for MLLMs in both the pre-training and adaptation phases.
    
    \item \textit{\textbf{Data evaluation methods and evaluation benchmarks.}} We outline commonly used methods to evaluate datasets as well as MLLMs evaluation benchmarks from a data-centric perspective.
    
    \item \textit{\textbf{Open issues and future directions.}} We discuss open issues in current research on data-centric LLMs and propose several future research directions.
\end{itemize}

\begin{figure}[!t]
\begin{center}
\centerline{\includegraphics[width=0.8\columnwidth]{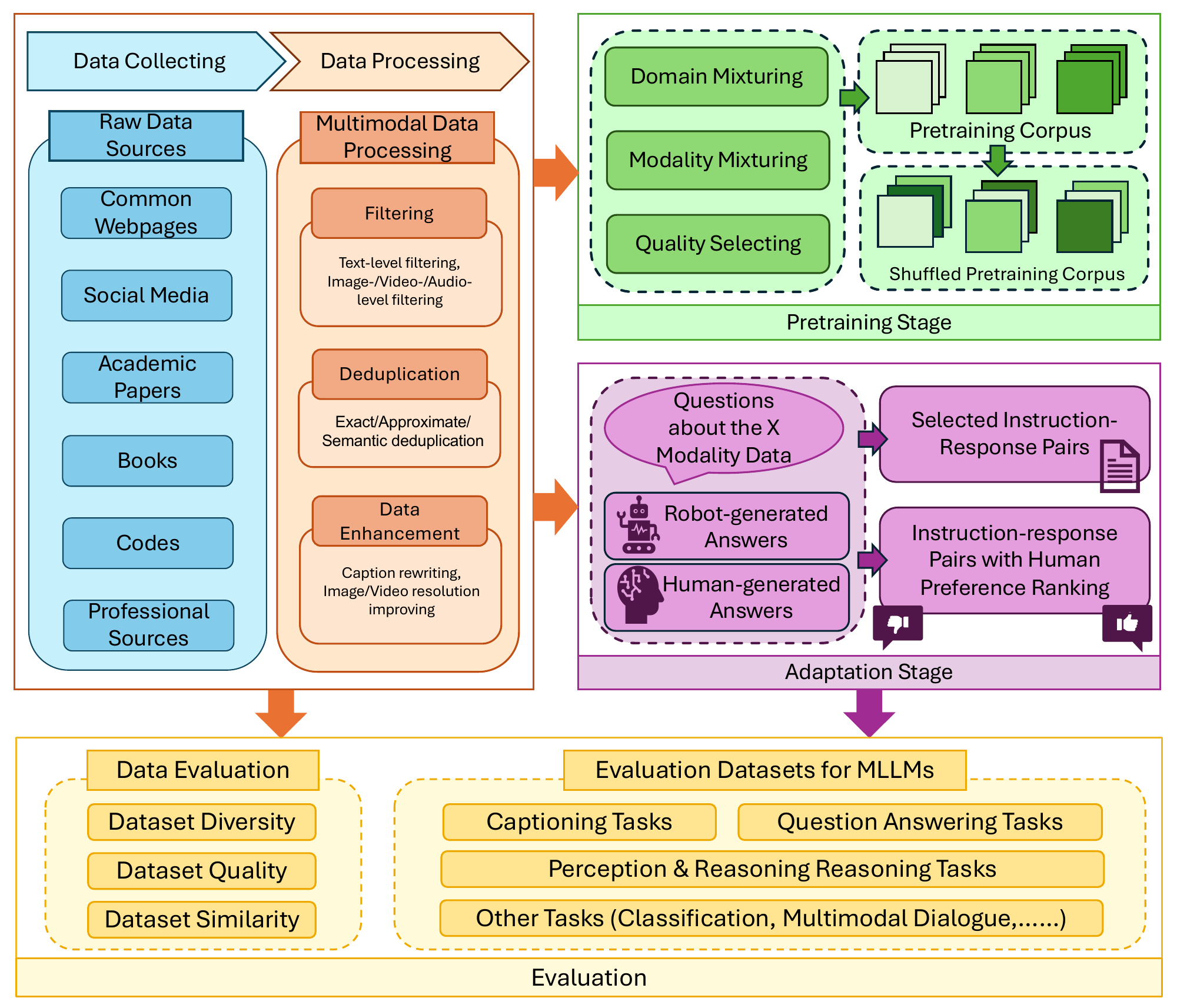}}
\caption{Overview of the data pipeline for MLLMs.}

\label{fig1:overview_graph}
\Description[Overview of data pipeline for MLLMs]{Overview of data pipeline for MLLMs}
\end{center}
\vspace{-4mm}
\end{figure}

The rest of this survey is organized as follows: Section~\ref{sec:overview} introduces the preliminaries for LLMs and  MLLMs, and discusses the motivations to analyze them from a data-centric perspective. Sections~\ref{sec:datapro} to~\ref{sec:adap} summarize the main stages of collecting, processing, and selecting data for MLLMs training.
Section~\ref{sec:eval} enumerates the data evaluation methods and existing evaluation datasets for MLLMs according to evaluation tasks. 
Section~\ref{sec:future} discusses open issues and highlights several future research directions in this field. Finally, we conclude the survey in Section~\ref{sec:conclu}. 

%% file: section/sec2_overview.tex
\section{backgrounds and categorization}\label{sec:overview}
\subsection{Large Language Models}
The field of natural language processing (NLP) has witnessed a remarkable evolution in language modeling techniques, culminating in the development of LLMs. The journey began with statistical methods \cite{jelinek1998statistical,gao2004introduction,rosenfeld2000two}, which laid the foundation for language modeling by capturing the probabilistic distributions of words and phrases. Subsequently, neural network approaches emerged~\cite{bengio2000neural,mikolov2010recurrent,kombrink2011recurrent}, leveraging the power of deep learning to learn more complex and abstract representations of language. However, such models were typically designed for specific tasks and encountered challenges related to scalability and generalization.

A significant breakthrough came with the introduction of pre-trained language models, which aimed to capture general language representations by training on vast amounts of unlabeled text data~\cite{kenton2019bert,radfordimproving}. Building upon these advancements, LLMs have taken the concept of pre-training to an unprecedented scale, with models containing billions of parameters and trained on massive corpora spanning hundreds of gigabytes to terabytes of text data~\cite{brown2020language,raffel2020exploring}.

LLMs exhibit several notable properties that distinguish them from previous language models. One fundamental characteristic is the scaling law, which describes how the performance of large language models improves as they scale in terms of model size, training data, and computational resources~\cite{kaplan2020scaling,hoffmann2022training}. This power-law relationship suggests that larger models trained on more data tend to capture more complex patterns and generalize better to new tasks. Another intriguing property of LLMs is the emergence of abilities that were not explicitly trained for, often referred to as emergent abilities~\cite{wei2022emergent}. This ability suggests that LLMs can capture and leverage complex linguistic patterns and knowledge from the pre-training data, enabling them to perform tasks beyond their original training objective. Furthermore, LLMs have demonstrated the capability of in-context learning~\cite{brown2020language}, where they can perform tasks based on a few examples provided in the input prompt, without the need for explicit fine-tuning. This highlights the models' ability to rapidly adapt to new tasks and generalize from limited examples.

The evolution of LLMs typically involves three key stages: training, adaptation, and evaluation~\cite{zhao2023survey}. The training stage focuses on learning general language representations from large-scale unlabeled corpora, capturing the underlying patterns and structures of natural language. The adaptation stage involves guiding the pre-trained model to specific tasks and human preference, often through supervised finetuning and human preference alignment~\cite{raffel2020exploring,ouyang2022training}. This step is crucial for optimizing the model's performance on the target application and ensuring its outputs are aligned with the desired objectives and human preference. Finally, the evaluation stage involves assessing the performance of the adapted model through various metrics and benchmarks to ensure that it meets the required standards and performs effectively in real-world scenarios.




\subsection{Multimodal Large Language Models}
Multimodal large language models (MLLMs) extend the capabilities of traditional LLMs by leveraging their comprehension, generation, and task-solving abilities across different modalities, including natural language (text and audio) and visual information (videos, images, and 3D models). The development of MLLMs is driven by the need to address real-world problems that often involve multiple modalities. For instance, in robotics, MLLM can process visual input from cameras, interpret natural language instructions, and analyze structured data from sensors to perform complex tasks~\cite{zeng2023large}. Similarly, in healthcare, MLLMs can analyze medical images, process electronic health records, and interpret patient-doctor conversations to aid in diagnosis and treatment planning.

Multimodal large language models (MLLMs) typically consist of three primary components: a modality encoder, a projector, and a large language model (LLM). The modality encoder is responsible for encoding data from different modalities, the projector aligns the data from various modalities with the LLM, and the LLM serves as the backbone, providing comprehension capabilities. By integrating LLMs with multimodal encoders and projectors, MLLMs can process various types of information, demonstrating strong understanding and analytical capabilities to address downstream tasks across various modalities~\cite{wu2023multimodal}.



MLLMs represent a significant advancement in AI by extending the capabilities of traditional LLMs to process and analyze multiple modalities. MLLMs can solve complex real-world problems that require understanding and reasoning across different types of information. As research in this field progresses, we can expect MLLMs to play an increasingly important role in various domains, from robotics and healthcare to education and entertainment.

\tikzset{
  FARROW/.style={arrows={-{Latex[length=1.25mm, width=1.mm]}}, }, 
  U/.style = {circle, draw=melon!400, fill=melon, minimum width=1.4em, align=center, inner sep=0, outer sep=0},
  I/.style = {circle, draw=tea_green!400, fill=tea_green, minimum width=1.4em, align=center, inner sep=0, outer sep=0},
  cate/.style = {rectangle, draw, minimum width=8em, minimum height=2em, align=center, rounded corners=3}, 
  cate2/.style = {rectangle, draw, minimum width=33em, minimum height=2em, text width=33em, align=left, rounded corners=3,anchor=west},
  cate3/.style = {rectangle, draw, minimum width=43em, text width=43em, minimum height=2em, align=left, rounded corners=3},
  encoder/.style = {rectangle, fill=Madang!82, minimum width=10em, minimum height=3em, align=center, rounded corners=3},
}

\begin{figure}
    \centering
    \resizebox{0.9\linewidth}{!}{
    \begin{tikzpicture}
    
    \node [cate, rotate=90, align=left] (n1) at (0, -1.5) {\textbf{Data-Centric MLLMs}};

    \node [cate,fill=orange!20, anchor=west] (n20) at (1,8) {Data Collecting\\and Processing};
    \draw[] (n1.south) -- (0.5,-1.5) -- (0.5,8) -- (n20.west);
    
    \node [cate, fill=cyan!20, anchor=west] (n201) at (4.5,11)  {Data Collecting\\Sources};
    \draw[] (n20.east) -- (4,8) -- (4,11) -- (n201.west);

    \node [cate, fill=cyan!20, anchor=west] (n2001) at (8,13.55)  {Common\\Webpages};
    \draw[] (n201.east) -- (7.5, 11)-- (7.5, 13.55)-- (n2001.west);
    \node [cate2, fill=cyan!20, anchor=west] (n20001) at (11.5,13.55)  {CommonCrawl Project~\cite{raffel2020exploring}, Self-Crawled Data: General Approches~\cite{sharma2018conceptual,changpinyo2021conceptual}, Focused Approaches (Wikipedia~\cite{gao2020pile,srinivasan2021wit,schamoni2018dataset}, Flickr~\cite{young2014image}, FreeSound~\cite{mei2023wavcaps, fonseca2022fsd50k}), Querying search engines-based approaches~\cite{wu2017ai,gu2022wukong}};
    \draw[] (n2001.east) --  (n20001.west);

    \node [cate, fill=cyan!20, anchor=west] (n2002) at (8,12.3)  {Social Media};
    \draw[] (n201.east) -- (7.5, 11)-- (7.5,12.3)--  (n2002.west);
    \node [cate2, fill=cyan!20, anchor=west] (n20002) at (11.5,12.3)  {Stack Exchange~\cite{cerebras2023slimpajama,chen2023fully}, Reddit~\cite{baumgartner2020pushshift, desai2021redcaps, overbay2023mredditsum}, YouTube~\cite{miech2019howto100m, zhou2017automatic}, X~\cite{barbieri2020tweeteval}};
    \draw[] (n2002.east) --  (n20002.west);

    \node [cate, fill=cyan!20, anchor=west] (n2003) at (8,11.4)  {Academic Papers};
    \draw[] (n201.east) -- (7.5, 11)-- (7.5, 11.4)--  (n2003.west);
    \node [cate2, fill=cyan!20, anchor=west] (n20003) at (11.5,11.4)  {ArXiv~\cite{together2023redpajama,gao2020pile,liu2023mmc}, S2ORC~\cite{lo2020s2orc,peS2o,soldaini2024dolma} };
    \draw[] (n2003.east) --  (n20003.west);
    
    \node [cate, fill=cyan!20, anchor=west] (n2004) at (8,10.35)  {Books};
    \draw[] (n201.east) -- (7.5, 11)-- (7.5, 10.35)--  (n2004.west);
    \node [cate2, fill=cyan!20, anchor=west] (n20004) at (11.5,10.35)  {Project Gutenberg~\cite{rae2019compressive}, Smashwords~\cite{zhu2015aligning}, Bibliotik~\cite{gao2020pile}, Amazon.com~\cite{mishra2019ocr}, Old Photo Books~\cite{okamoto2023constructing}};
    \draw[] (n2004.east) --  (n20004.west);

    \node [cate, fill=cyan!20, anchor=west] (n2005) at (8,9.3)  {Codes};
    \draw[] (n201.east) -- (7.5, 11)-- (7.5,9.3)--  (n2005.west);
    \node [cate2, fill=cyan!20, anchor=west] (n20005) at (11.5,9.3)  {BigQuery~\cite{nijkamp2022codegen}, GitHub~\cite{kocetkov2022stack,nijkamp2022codegen, gao2020pile}, The Stack~\cite{li2023starcoder}};
    \draw[] (n2005.east) --  (n20005.west);

    \node [cate, fill=cyan!20, anchor=west] (n2006) at (8,7.85)  {Professional\\Sources};
    \draw[] (n201.east) -- (7.5, 11)-- (7.5,7.85)--  (n2006.west);
    \node [cate2, fill=cyan!20, anchor=west] (n20006) at (11.5,7.85)  {Legal Domain: FreeLaw~\cite{gao2020pile}; Math Domain: Khan Academy, DeepMind~\cite{saxton2018analysing}; Medical Domain: Online Medical Websites~\cite{li2023huatuo26m,ben2019question}, Knowledge Bases~\cite{li2023huatuo26m,welbl2018constructing}, In-Hospital Database Systems~\cite{johnson2023mimic,johnson2019mimiccxrjpg,bustos2020padchest}; Financial Domain: Financial Platforms and Forums~\cite{ok2023fintree, lu2023bbt}};
    \draw[] (n2006.east) --  (n20006.west);

    \node [cate, fill=orange!20, anchor=west] (n202) at (4.5,5.05)  {Data Processing\\ Methods};
    \draw[] (n20.east) -- (4,8) -- (4,5.05) -- (n202.west);

    \node [cate, fill=orange!20, anchor=west] (n2021) at (8,6.25)  {Filtering};
    \draw[] (n202.east) -- (7.5, 5.05)-- (7.5,6.25)--  (n2021.west);
    \node [cate2, fill=orange!20, anchor=west] (n20211) at (11.5,6.25)  {Textual Filtering~\cite{joulin2016fasttext, grave2018learning, raffel2020exploring,raffel2020exploring, yuan2021wudaocorpora, gunasekar2023textbooks}, Image-Level Filtering~\cite{zhu2024multimodal, guo2021sample, yang2022study}, Video-Level Filtering~\cite{wang2023internvid,blattmann2023stable,farneback2003two,baek2019character}};
    \draw[] (n2021.east) --  (n20211.west);

    \node [cate, fill=orange!20, anchor=west] (n2022) at (8,5.05)  {Deduplication};
    \draw[] (n202.east) -- (7.5, 5.05)-- (7.5, 5.05)--  (n2022.west);
    \node [cate2, fill=orange!20, anchor=west] (n20212) at (11.5,5.05)  {Exact Duplication~\cite{raffel2020exploring,suarez2019asynchronous, lee2022deduplicating}, Approximate Duplication~\cite{together2023redpajama,lee2022deduplicating,penedo2023refinedweb}, Sentence-Level and Document-Level Semantic Duplication~\cite{reimers2019sentence, song2020mpnet, silcock2022noise, abbas2023semdedup}};
    \draw[] (n2022.east) --  (n20212.west);

    \node [cate, fill=orange!20, anchor=west] (n2021) at (8,4)  {Data\\Enhancement};
    \draw[] (n202.east) -- (7.5, 5.05)-- (7.5, 4)--  (n2021.west);
    \node [cate2, fill=orange!20, anchor=west] (n20211) at (11.5,4)  {Enhance X Modality Data~\cite{li2023monkey, luo2024feast, wei2023vary}, Enhance Text Data~\cite{Nguyen2023ImprovingMD, fan2024improving, Lai2023VeCLIPIC, chen2023sharegpt4v}};
    \draw[] (n2021.east) --  (n20211.west);

    \node [cate, fill=orange!20, anchor=west] (n203) at (4.5, 0.8)  {Commonly-Used \\Datasets};
    \draw[] (n20.east) -- (4,8) -- (4,0.8) -- (n203.west);

    \node [cate, fill=orange!20, anchor=west] (n2031) at (8,2.75)  {Image datasets};
    \draw[] (n203.east) -- (7.5, 0.8)-- (7.5, 2.75)--  (n2031.west);
    \node [cate2, fill=orange!20, anchor=west] (n20311) at (11.5,2.75)  {Image-Caption Datasets~\cite{schuhmann2021laion, schuhmann2022laion, schuhmann2022laion}, Content Descriptive Image-Caption Dataset~\cite{chen2015microsoft, young2014image, chen2022pali}, Interleaved Image-Text Documents~\cite{alayrac2022flamingo, zhu2024multimodal, laurenccon2024obelics}, Visual Question Answer Image-Caption Datasets~\cite{goyal2019making, zhu2016visual7w, biten2019scene}};
    \draw[] (n2031.east) --  (n20311.west);

    \node [cate, fill=orange!20, anchor=west] (n2031) at (8,1.35)  {Video datasets};
    \draw[] (n203.east) -- (7.5, 0.8)-- (7.5, 1.35)--  (n2031.west);
    \node [cate2, fill=orange!20, anchor=west] (n20311) at (11.5,1.35)  {Video Captioning Datasets~\cite{xu2016msr, bain2021frozen, alayrac2022flamingo}, Video Question Answering Datasets~\cite{chen2011collecting, xu2016msr, jang2017tgif}, Interleaved Video-Text Dataset~\cite{sun2023generative}};
    \draw[] (n2031.east) --  (n20311.west);

    \node [cate, fill=orange!20, anchor=west] (n2031) at (8,0.3)  {Audio datasets};
    \draw[] (n203.east) -- (7.5,0.8)-- (7.5, 0.3)--  (n2031.west);
    \node [cate2, fill=orange!20, anchor=west] (n20311) at (11.5,0.3)  {Audio Captioning Datasets~\cite{lipping2022clotho, li2022learning}, Question Answering Datasets~\cite{kim2019audiocaps,drossos2020clotho}};
    \draw[] (n2031.east) --  (n20311.west);

    \node [cate, fill=orange!20, anchor=west] (n2031) at (8,-0.6)  {3D datasets};
    \draw[] (n203.east) -- (7.5, 0.8)-- (7.5, -0.6)--  (n2031.west);
    \node [cate2, fill=orange!20, anchor=west] (n20311) at (11.5,-0.6)  {3D Captioning Datasets~\cite{dai2017scannet, armeni2017joint, Structured3D}};
    \draw[] (n2031.east) --  (n20311.west);

    \node [cate, fill=green!20, anchor=west] (n21) at (1,-2.5) {Data-Centric\\Pre-training};
    \draw[] (n1.south) -- (0.5,-1.5) -- (0.5,-2.5) -- (n21.west);
    
    \node [cate, fill=green!20, anchor=west] (n211) at (4.5, -1.5)  {Domain Mixture};
    \draw[] (n21.east) -- (4,-2.5) -- (4,-1.5) -- (n211.west);
    \node [cate3, fill=green!20, anchor=west] (n2111) at (8,-1.5)  {Pure Text Domain Mixture~\cite{xie2024doremi, xie2024data, du2022glam}, Multimodal Domain Mixture~\cite{xu2023youku,chen2024vast, song2023moviechat, chen2023videollm,wang2023internvid}};
    \draw[] (n211.east) --  (n2111.west);

    \node [cate, fill=green!20, anchor=west] (n212) at (4.5, -2.5)  {Modality Mixture};
    \draw[] (n21.east) -- (4,-2.5) -- (4,-2.5) -- (n212.west);
    \node [cate3, fill=green!20, anchor=west] (n2121) at (8,-2.5)  {Image Caption Data Mixture~\cite{mckinzie2024mm1, lai2023scarcity}, Image and Video Caption Data Mixture~\cite{lin2023video,zhang2023video,luo2023valley, jin2023chat, li2023llama, han2023autoad,li2023videochat,li2023mvbench,wang2023internvid,chen2023valor}};
    \draw[] (n212.east) --  (n2121.west);

    \node [cate, fill=green!20, anchor=west] (n213) at (4.5, -3.65)  {Quality Selection};
    \draw[] (n21.east) -- (4,-2.5) -- (4,-3.65) -- (n213.west);
    \node [cate3, fill=green!20, anchor=west] (n2131) at (8,-3.65)  {Active Learning-Based Selection~\cite{Xu2023CiTCI}, Selection Before Training~\cite{gadre2024datacomp}, Distribution-Agnostic Selection~\cite{gadre2024datacomp, wang2024finetuned, mahmoud2024sieve}, Distribution-Aware Selection~\cite{wang2024variance, gadre2024datacomp}};
    \draw[] (n213.east) --  (n2131.west);

    \node [cate, fill=violet!20, anchor=west] (n22) at (1,-8) {Data-Centric \\ Adaptation};
    \draw[] (n1.south) -- (0.5,-1.5) -- (0.5,-8) -- (n22.west);

    \node [cate, fill=violet!20, anchor=west] (n221) at (4.5,-6.15)  {Supervised\\ Finetuning};
    \draw[] (n22.east) -- (4,-8) -- (4,-6.15) -- (n221.west);
    
    \node [cate, fill=violet!20, anchor=west] (n2211) at (8,-4.68)  {Data Processing};
    \draw[] (n221.east) -- (7.5, -6.15)-- (7.5, -4.68)--  (n2211.west);
    \node [cate2, fill=violet!20, anchor=west] (n22111) at (11.5,-4.68)  {SFT Data Processing~\cite{zhu2023minigpt4, chen2023sharegpt4v}};
    \draw[] (n2211.east) --  (n22111.west);

    \node [cate, fill=violet!20, anchor=west] (n2212) at (8,-6.15)  {Data Generation};
    \draw[] (n221.east) -- (7.5,-6.15)-- (7.5, -6.15)--  (n2212.west);
    \node [cate2, fill=violet!20, anchor=west] (n22121) at (11.5,-6.15)  {Captioning Instruction-Response Datasets Generation~\cite{zhu2023minigpt4, xu2024llava, openai2023gpt}, Question Answer Instruction-Response Datasets Generation~\cite{liu2023improved, singh2019towards, li2023m}, Reasoning Instruction-Response Datasets Generation~\cite{chen2023shikra, chen2023position, chen2023minigptv2}, Other Instruction-Response Datasets Generation~\cite{li2023m,panagopoulou2023x,dai2024instructblip}};
    \draw[] (n2212.east) --  (n22121.west);

    \node [cate, fill=violet!20, anchor=west] (n2212) at (8,-7.95)  {Data Selection};
    \draw[] (n221.east) -- (7.5, -6.15)-- (7.5, -7.95)--  (n2212.west);
    \node [cate2, fill=violet!20, anchor=west] (n22121) at (11.5,-7.95)  {Coreset-Based Selection~\cite{chen2023maybe, das2023deft, sorscher2022beyond}, LLMs-Based Selection~\cite{chen2023alpagasus, xu2023rethinking, liu2023makes}, gradient-Based Selection~\cite{paul2021deep, attendu2023nlu, xia2024less}, Self-Instruction-Based Selection Methods~\cite{li2023quantity,li2023one,kung2023active,liu2024selectit}};
    \draw[] (n2212.east) --  (n22121.west);

    \node [cate, fill=violet!20, anchor=west] (n222) at (4.5,-9.2)  {Alignment};
    \draw[] (n22.east) -- (4,-8) -- (4,-9.2) -- (n222.west);
    \node [cate3, fill=violet!20, anchor=west] (n2221) at (8,-9.2)  {Human Preference Alignment: Pure Text~\cite{ouyang2022training, li2023llama}, Multimodal~\cite{liu2023visual, sun2023aligning, chen2023dress, ouyang2022training}};
    \draw[] (n222.east) -- (n2221.west);

    \node [cate, fill=yellow!20, anchor=west] (n23) at (1,-12.35){Data-Centric\\Evaluation};
    \draw[] (n1.south) -- (0.5,-1.5) -- (0.5,-12.35) -- (n23.west);

    \node [cate, fill=yellow!20, anchor=west] (n231) at (4.5,-10.1)  {Data Evaluation};
    \draw[] (n23.east) -- (4,-12.35) -- (4,-10.1) -- (n231.west);
    \node [cate3, fill=yellow!20, anchor=west] (n2311) at (8,-10.1)  {Dataset Diversity~\cite{heusel2017gans,friedman2023vendi}; Dataset Quality~\cite{honovich2022true,rohrbach2018object,jing2023faithscore}; Dataset Similarity~\cite{sun2016deep,jiang2022transferability,pillutla2023mauve}};
    \draw[] (n231.east) -- (n2311.west);

    \node [cate, fill=yellow!20, anchor=west] (n232) at (4.5,-13)  {Evaluation \\Datasets};
    \draw[] (n23.east) -- (4,-12.35) -- (4,-13) -- (n232.west);
    
    \node [cate, fill=yellow!20, anchor=west] (n2321) at (8,-11.15)  {Captioning\\Tasks};
    \draw[] (n232.east) -- (7.5, -13)-- (7.5, -11.15)--  (n2321.west);
    \node [cate2, fill=yellow!20, anchor=west] (n23211) at (11.5,-11.15)  {Image Captioning Datasets~\cite{goyal2019making, hudson2019gqa, marino2019ok}, Video Captioning Datasets~\cite{xu2017video, chen2011collecting, yu2019activitynet}, Audio Captioning Datasets~\cite{lipping2022clotho, li2022learning}};
    \draw[] (n2321.east) --  (n23211.west);

    \node [cate, fill=yellow!20, anchor=west] (n2322) at (8,-12.35)  {Question\\Answering Tasks};
    \draw[] (n232.east) -- (7.5, -13)-- (7.5, -12.35)--  (n2322.west);
    \node [cate2, fill=yellow!20, anchor=west] (n23221) at (11.5,-12.35)  {Image Question Answering Datasets~\cite{chen2015microsoft, young2014image, karpathy2015deep}, Video Question Answering Datasets~\cite{chen2011collecting, venugopalan2015translating, xu2016msr}, Audio Question Answering Datasets~\cite{kim2019audiocaps,drossos2020clotho}};
    \draw[] (n2322.east) --  (n23221.west);

    \node [cate, fill=yellow!20, anchor=west] (n2323) at (8,-13.55)  {Perception \&\\Reasoning Tasks};
    \draw[] (n232.east) -- (7.5, -13)-- (7.5, -13.55)--  (n2323.west);
    \node [cate2, fill=yellow!20, anchor=west] (n23231) at (11.5,-13.55)  {Perception and Reasoning Datasets~\cite{singh2019towards, masry2022chartqa, kembhavi2016diagram}, Spatial Reasoning Datasets~\cite{liu2023visual, kazemzadeh2014referitgame, mao2016generation}};
    \draw[] (n2323.east) --  (n23231.west);

    \node [cate, fill=yellow!20, anchor=west] (n2324) at (8,-14.75)  {Other Noteworthy\\Tasks};
    \draw[] (n232.east) -- (7.5, -13)-- (7.5, -14.75)--  (n2324.west);
    \node [cate2, fill=yellow!20, anchor=west] (n23241) at (11.5,-14.75)  {Classification Datasets~\cite{kiela2020hateful, deng2009imagenet, piczak2015esc, wu20153d, goyal2017something}, Multimodal Dialogue Datasets~\cite{bai2023touchstone}};
    \draw[] (n2324.east) --  (n23241.west);

    \end{tikzpicture}}
    \caption{Overview of Data-Centric MLLMs}
    \label{fig2:overview_table}
    \Description[Overview of Data-Centric MLLMs]{Categorization of Data-Centric MLLMs work}
\end{figure}
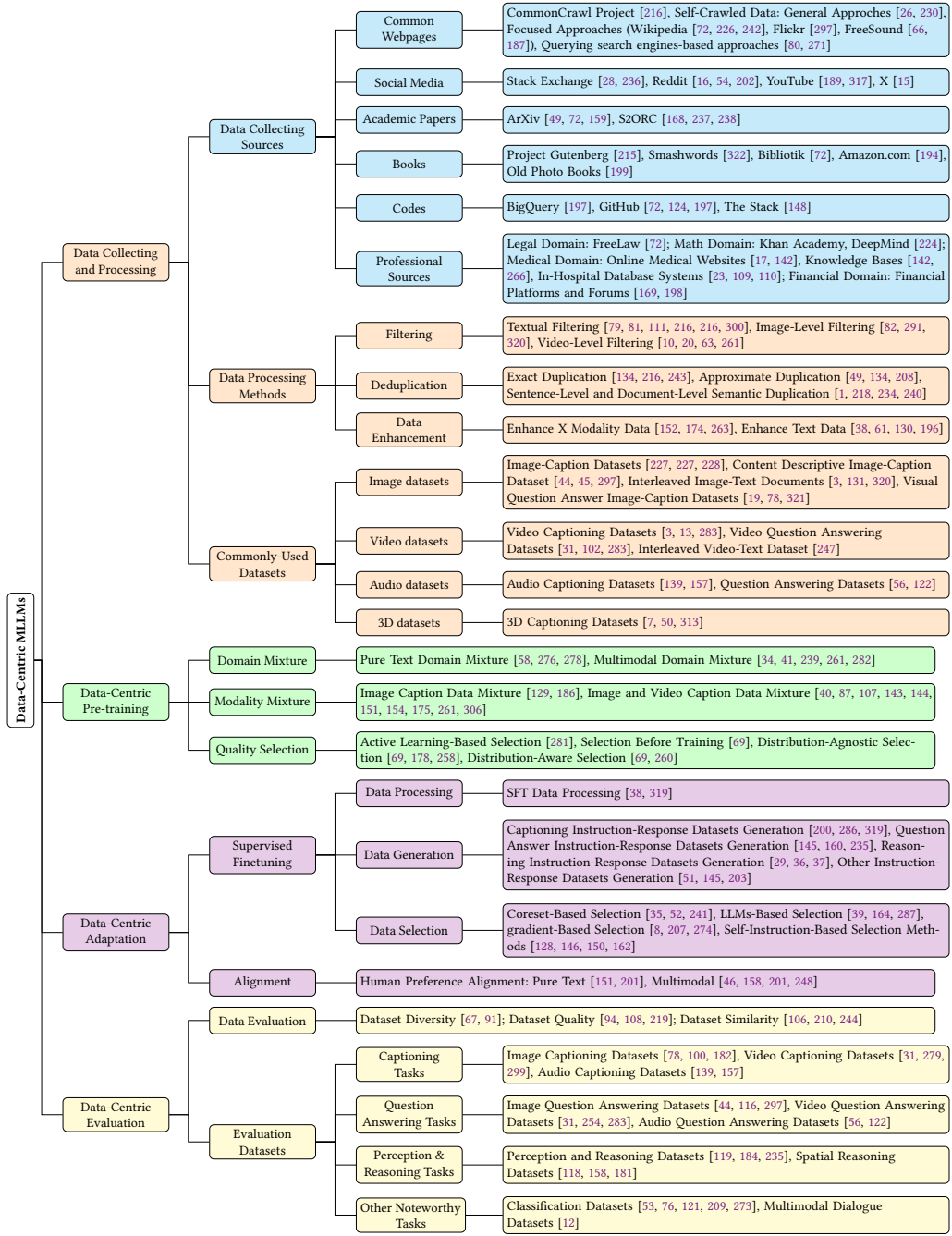

\subsection{Data-Centric AI and Why Data-Centric MLLMs}
The field of artificial intelligence (AI) has experienced a paradigm shift towards data-centric approaches in recent years~\cite{zha2023data}. Data-centric AI emphasizes the critical role of data quality, diversity, and representativeness in building robust and effective AI systems. This shift acknowledges that the performance of AI models is heavily based on the quality and characteristics of the data used for training, rather than solely focusing on algorithmic improvements.

In the context of LLMs and MLLMs, data play a pivotal role in determining their capabilities and limitations. LLMs, such as GPT-3~\cite{brown2020language} and BERT~\cite{kenton2019bert}, are trained on vast amounts of textual data to learn the intricacies of language and generate coherent, contextually relevant outputs. Similarly, MLLMs, which integrate multiple modalities such as text, images, and speech, require diverse and well-aligned datasets in different modalities to learn meaningful cross-modal representations~\cite{gadre2024datacomp}. The quality, diversity, and representativeness of the training data directly impact the models' ability to understand and generate language, as well as their capacity to reason and perform tasks across multiple modalities~\cite{llama3repo, liu2023improved, chen2023sharegpt4v}.

Analyzing LLMs and MLLMs from a data-centric perspective offers several advantages. First, it enables researchers to identify and address potential biases and limitations in the training data that can propagate to the models' outputs. By carefully curating and augmenting datasets to ensure diversity and representativeness, researchers can mitigate biases and improve the fairness and generalizability of the models~\cite{chen2024mllm}. Second, a data-centric approach allows for a deeper understanding of the models' capabilities and limitations based on the characteristics of the training data~\cite{mckinzie2024mm1}. By systematically varying data properties and evaluating model performance, researchers can gain insights into the specific data attributes that contribute to the models' success or failure in various tasks. This understanding can guide the development of more efficient and effective data collection and curation strategies. Moreover, adopting a data-centric perspective in the development of LLMs and MLLMs opens up opportunities for data-efficient training and adaptation~\cite{liu2024visual, liu2023improved, chen2023sharegpt4v}. Additionally, data-centric approaches facilitate the development of more interpretable and explainable models by linking model behaviors to specific data characteristics.


\subsection{Categories of Data-Centric MLLMs}
In this survey, we review previous work on MLLMs from a data-centric perspective. Our taxonomy is primarily based on different stages of MLLM development, as illustrated in Figure~\ref{fig2:overview_table}. Specifically, we organize our article around a data pipeline, shown in Figure~\ref{fig1:overview_graph}, where the data pipeline for MLLMs comprises three stages: pre-processing (including data collection and data processing in Section~\ref{sec:datapro}), pre-training data processing in Section~\ref{sec:pre}), and adaptation data processing in Section~\ref{sec:adap}. 
Along this pipeline, we discuss data-related model performance and analyze how dataset curation affects model performance. 
During the pre-processing stage, large-scale multimodal data are collected from various sources and constructed into datasets. Processing steps such as filtering, deduplication, and data enhancement are applied to improve data quality. In the pre-training stage, data selection, domain mixing and modality mixing are performed to choose appropriate data for model training. In the adaptation stage, we focus on data processing and generation for supervised fine-tuning and human preference alignment.
Furthermore, we discuss data evaluation metrics and evaluation datasets for MLLMs in Section~\ref{sec:eval}, examining their construction and main characteristics. Finally, we suggest possible future directions for further research on data-centric MLLMs in Section~\ref{sec:future}.

%% file: section/sec3_datapro.tex
\section{Data collecting and processing}\label{sec:datapro}
The first and fundamental step in training MLLMs is to collect and process sufficient data from various sources. In this section, we first introduce the sources for data collection. Next, we summarize the processing steps for multimodal data, including commonly used filtering, deduplication, and data enhancement methods. Finally, we provide an overview of commonly used datasets for MLLMs. 

\subsection{Data Collecting Sources}\label{pre:collect}
We first introduce the sources where researchers typically find raw data for MLLM pre-training. Commonly used data sources can be divided into six categories: common webpages, social media, academic papers, books, code repositories, and professional sources. Data from these different sources have various characteristics, each contributing uniquely to improving different abilities of the models.


\paragraph{Common Webpages.} Webpages are the main source for collecting large-scale training corpus. CommonCrawl project serves as the most commonly used start-point for large-scale webpages. It is instrumental in generating large-scale pre-training corpora like C4~\cite{raffel2020exploring} for LLM pre-training. Beyond textual content, CommonCrawl's vast archive of web pages, which includes numerous image-text pairs, has also become a vital resource for constructing multimodal pre-training datasets such as LAION-5B~\cite{schuhmann2022laion}.

Apart from CommonCrawl, several works focus on selecting specific webpages and crawling data independently. For textual datasets, initiatives like WuDaoCorpora~\cite{yuan2021wudaocorpora} crawl thousands of web pages to construct vast textual corpora. In the multimodal dataset sector, there are three main self-crawling approaches: general approach, focused approach on specific platforms, and querying search engines for images. Examples include AI Challenger Captions~\cite{wu2017ai} and Wukong~\cite{gu2022wukong}. General approaches, such as CC3M~\cite{sharma2018conceptual} and CC12M~\cite{changpinyo2021conceptual}, involve crawling billions of web pages without targeting specific platforms. Focused approaches, on the other hand, involve selecting specific webpages to crawl. For example, Wikipedia is a commonly used resource that serves as a rich and accurate source for text-based single-modal pre-training datasets like the Pile~\cite{gao2020pile}. It also plays a crucial role in creating multimodal pre-training datasets such as WIT~\cite{srinivasan2021wit} and WikiCaps~\cite{schamoni2018dataset}.
For the vision modality, Flickr~\cite{young2014image} serves as an important source, containing photos and videos shared by online users. This contributes to image-text datasets such as Flickr30k~\cite{young2014image}. For the audio modality, platforms such as BBC Sound Effects, FreeSound, and SoundBible are commonly used, contributing to datasets such as WavCaps~\cite{mei2023wavcaps} and FSD50K~\cite{fonseca2022fsd50k}.

\paragraph{Social Media.} 
Social media platforms significantly enhance the training datasets for MLLMs by providing real-time, varied, and colloquial text data, as well as multimodal content that captures human expression and interaction. However, questions regarding the ownership and copyright of social media content are on the rise. MLLM researchers need to be aware of these issues and understand the rights and licensing of the data sources they use.
Stack Exchange offers high-quality textual and visual content, hosting extensive forums with rich dialogues and images. In addition to Stack Exchange, Reddit offers user-generated content with discussions and voting on a wide array of topics, contributing to both textual and visual content. For video data, YouTube is a standout platform, with a vast amount of videos uploaded every minute. X, formerly known as Twitter, offers great amount of sharing text messages, images, audio, and videos.


\paragraph{Academic Papers.}
Academic papers offer high-quality, authoritative content, enriching pre-training datasets with specialized knowledge and formal language, which is critical for developing language models with expertise in academic and professional domains. Academic papers are primarily utilized as sources for text-based pre-training datasets. arXiv offers a vast array of prepublished scientific research for training specialized language models. This is evidenced by its role in creating datasets such as RedPajama-Data-1T~\cite{together2023redpajama}, Pile-arXiv~\cite{gao2020pile}, and image-text datasets such as MMC~\cite{liu2023mmc}. Another academic platform semantic scholars contribute to the S2ORC~\cite{lo2020s2orc} corpus, which contains 81.1 million academic papers, offering metadata, abstracts, references, and full texts for 8.1 million open-access works. 

\paragraph{Books.}
Books provide a rich source of high-quality textual content for language model training and serve as fertile ground for multimodal pre-training datasets. They offer diverse visual data from book covers and illustrations, enriching models with a blend of literary depth and visual context.

Project Gutenberg acts as a pivotal source for language modeling and analysis with its extensive library of more than 70,000 free eBooks, fueling datasets such as PG-19~\cite{rae2019compressive}. Smashwords contributes to the well-known BookCorpus~\cite{zhu2015aligning} dataset, which encompasses a diverse collection of 11,038 self-published novels that span genres such as romance, science fiction, and fantasy. This extensive corpus has been instrumental in training landmark models such as GPT~\cite{radfordimproving} and BERT~\cite{kenton2019bert}. Bibliotik that consists of a mix of fiction and non-fiction books contributes to Books3~\cite{gao2020pile}, part of the Pile dataset, which contains about 197,000 books. This dataset is essential for context modeling and narrative research, accounting for 2.1\% of the RedPajama-Data-1T dataset.

Meanwhile, books also serve as an important source for constructing multimodal datasets. Book covers hold vast multimodal information, combining visual elements with metadata such as author names, titles, and genres. This fusion of data supports tasks like genre classification and visual question-answering. The OCR-VQA dataset~\cite{mishra2019ocr}, sourced from Amazon.com, exemplifies how book covers can be leveraged to build rich pre-training datasets for MLLMs, demonstrating the value of covers in enhancing model understanding of visual and textual content. Old photo books offer a unique advantage for constructing multimodal pre-training datasets by providing historical visual and textual contexts, enriching models' understanding of past societies and landscapes. 
For example, previous work collected 9,516 image-text pairs from 175 Japanese old photo books~\cite{okamoto2023constructing}.


\paragraph{Domain-Specific Sources.}
To enhance the performance of MLLMs in specific domains, general pre-training is supplemented with incremental, domain-specific pre-training. In the legal domain, pre-training data is primarily textual. FreeLaw stands out by offering free access to judgments from India's higher courts, exemplified by the Pile-FreeLaw~\cite{gao2020pile} dataset, which structures these judgments for machine learning. In the math domain, textual datasets primarily originate from Khan Academy's exercises for foundational concepts and DeepMind's Pile-DeepMind Mathematics~\cite{saxton2018analysing} for advanced, algorithmically generated problems. Additionally, DVQA~\cite{kafle2018dvqa} introduces a multimodal aspect by combining bar charts with question-answer pairs generated through Matplotlib. In the medical domain, pre-training data comes from online medical websites, knowledge bases, and in-hospital database systems. Online medical websites such as Qianwen Health and PubMed contribute to datasets like Huatuo-26M~\cite{li2023huatuo26m} and MedQuAD~\cite{ben2019question}. Knowledge bases like Wikipedia also contribute to Huatuo-26M~\cite{li2023huatuo26m} and MedHop~\cite{welbl2018constructing}. In-hospital database systems, including electronic health record systems (EHR) and ICU-specific clinical information systems, are used in MIMIC-IV~\cite{johnson2023mimic}. Multimodal datasets in radiology, such as MIMIC-CXR-JPG~\cite{johnson2019mimiccxrjpg} and PADCHEST~\cite{bustos2020padchest}, are generated from in-hospital radiology reports. In the financial domain, pre-training primarily relies on textual data sourced from both English and Chinese platforms. English contributions come from EDGAR and the SEC Financial Statement and Notes Data Sets, which enhance the FinTree project~\cite{ok2023fintree}. Chinese sources include major sites like Sina Finance and Tencent Finance, as well as Eastmoney for specific documents, and forums like Guba and Xueqiu, culminating in the comprehensive BBT-FinCorpus~\cite{lu2023bbt}. This assortment of sources underscores the rich textual foundation for financial model pre-training.

\subsection{Data Processing}\label{sec:pre:datapro}
\subsubsection{Filtering}
Data filtering in MLLM training is critical for enhancing model reliability and efficiency. Filtering out unwanted content is essential, as even minimal exposure to hate speech can negatively influence model behavior~\cite{luccioni2021s, gunasekar2023textbooks}. For multimodal datasets, previous work has focused on filtering out unwanted data from different modalities separately. 
Specifically, for X-Text datasets, they consider filtering data based on both the text and the X modality independently.

Textual filters primarily encompass language and content filtering. For language filtering, documents or sentences that fall below a certain threshold for a specific language are removed.
For English-only datasets, commonly used tools like Langdetect and FastText operate at the document level. For instance, C4~\cite{raffel2020exploring} utilizes Langdetect to filter out any non-English pages with a probability below 0.99.
Another renowned English language filter, FastText, is widely employed by datasets such as Dolma~\cite{soldaini2024dolma} and RefinedWeb~\cite{penedo2023refinedweb}. 
CLUECorpus2020~\cite{xu2020cluecorpus2020} adopts a sentence-level filtering approach, selecting sentences whose language type is Chinese if a language is mentioned.
For multilingual datasets, FastText, trained on Wikipedia data to classify 176 languages, remains a preferred choice, capable of processing 1,000 documents per second on a single CPU core. ROOTS~\cite{laurenccon2022bigscience} utilizes FastText for document-level language classification, resulting in a 1.6TB dataset containing 59 languages.
For code datasets, detection methods are more elementary, often based on file extensions. For instance, in Palm~\cite{chowdhery2023palm}, files are filtered based on filename extensions to restrict to one of 24 common programming languages, resulting in 196GB of source code.

Another type of textual filtering is content filtering, which includes removing toxic and distracting content. Toxic content filtering targets text that is deemed rude, disrespectful, or unreasonable, employing heuristic and rule-based methods. For instance, in C4~\cite{raffel2020exploring}, pages containing words in a predefined "Dirty, Naughty, Obscene, or Otherwise Bad Words" list are removed.
Distracting content includes short or incomplete sentences, "lorem ipsum" text, and useless identifiers like HTML, CSS, and JavaScript tags. Filtering methods usually involve multiple rule-based filters, such as punctuation-based segmentation in WudaoCorpora~\cite{yuan2021wudaocorpora}, or machine learning approaches, as seen in phi-1~\cite{gunasekar2023textbooks}, which use GPT-4 as an annotators and then train a high-quality content classifier based on the annotated results.

For image-level filtering, the most fundamental step is to remove images with excessively low resolution, as these images often fail to convey effective information. Additionally, it is necessary to filter out images with inappropriate aspect ratios, as such unconventional images frequently resemble banner-like advertisements~\cite{zhu2024multimodal}. Furthermore, similar to removing harmful content in text filtering, such as NSFW material, it is necessary to filter out such content from images. This can be achieved by training a binary classification model using appropriate datasets. Additionally, any potential occurrences of human faces or other sensitive elements in the images should be detected using a face detector and blurred accordingly~\cite{guo2021sample}. These filtering processes have a minimal effect on the overall model performance~\cite{yang2022study}.

For video-level filtering, previous work has leveraged image or image-text filtering methods to handle static content and applied specific techniques for dynamic content. Video filtering involves four main components: scene transition detection, video quality and integrity improvement, refinement and coherence evaluation, and modality completeness. For scene transition detection, algorithms are used to remove abrupt scene changes, ensuring a cleaner dataset free from disruptive transitions~\cite{wang2023internvid,blattmann2023stable}. To improve video quality and integrity, three aspects are considered: motion, textual information in the video, and resolution and frame rate. For motion analysis, dense optical flow~\cite{farneback2003two} is typically utilized to measure motion complexity and filter out static or repetitive sequences with limited educational value. 
To detect textual information in videos, OCR~\cite{baek2019character} is applied to identify and remove clips with substantial text, which can otherwise mislead the model's visual interpretation. 
The resolution and frame rate of training video datasets are standardized, enabling model to learn from uniformly structured data~\cite{wang2023internvid}.
In constructing video-text datasets, ensuring the relevance and clarity of annotations is paramount. 
Focused efforts are made to refine the textual content of video-text pairs, particularly subtitles, to ensure they are contextually meaningful and devoid of promotional or irrelevant material~\cite{xu2023youku}. 
Models such as CLIP are employed to evaluate the coherence between video frames and the accompanying text~\cite{chen2024vast,xu2023youku,wang2023internvid}, ensuring that annotations accurately reflect the video content. 
In some studies, clips missing any modalities—vision, audio, or subtitles—are excluded to meet the requirements for comprehensive understanding across all modalities~\cite{chen2024vast}.

\subsubsection{Deduplications}
Previous work has found there are great amount of duplicate data in various training datasets. 
For example, the BOOK Corpus~\cite{zhu2015aligning} contains thousands of duplicated books~\cite{bandy2021addressing}. 
The commonly used curated web crawl dataset C4~\cite{raffel2020exploring} contains a single 61 word English sentence that repeated more than 6 hundred thousands times~\cite{lee2022deduplicating}. 
More repeated data in training datasets can increase the rate of emitting memorized training data verbatim~\cite{carlini2022quantifying,kandpal2022deduplicating}, and deduplication of training datasets can prevent the memorization problem, thus alleviating privacy concerns~\cite{kandpal2022deduplicating}. 
Moreover, duplication of training data can also cause the performance degradation of the pre-trained models~\cite{hernandez2022scaling}. 
Training on deduplicated datasets can save the training cost while does not hurt the model perplexity~\cite{lee2022deduplicating}.
    
Existing deduplication methods consider exact duplication, approximate duplication and semantic duplication in sentence-level (or sequence-level) and document-level. 
\textbf{Exact deduplication} is the most simple way to remove duplication. 
For sentence or sequence level, exact deduplication consider to remove exact string matching between sentence (or sequence)~\cite{raffel2020exploring,suarez2019asynchronous}. 
To improve computational and memory efficiency, Suffix Arrays~\cite{manber1993suffix, lee2022deduplicating} and Bloom Filters~\cite{together2023redpajama, soldaini2024dolma} are used to achieve parallelized linear time complexity. 
For document level exact deduplication, URL deduplication is considered to remove exactly same web pages~\cite{penedo2023refinedweb}.
\textbf{Approximate deduplication} methods mainly focus on document-level duplicates. 
Usually, Locality Sensitive Hashing (LSH)-based MinHash~\cite{broder1997resemblance} or SimHash~\cite{charikar2002similarity} methods are used to remove approximately duplicate documents~\cite{together2023redpajama,lee2022deduplicating,penedo2023refinedweb}. 
These methods can achieve document-level deduplication with linear time and space complexity of the document number and can be implemented in a highly distributed setting~\cite{lee2022deduplicating}. Apart from hashing-based approximate deduplication, recently there are some work consider to leverage pre-trained foundation models as semantic embedding metric for document-level embedding~\cite{silcock2022noise,kaddour2023minipile,abbas2023semdedup,tirumala2023d4}. These \textbf{Semantic deduplication} methods consider using Sentence-BERT~\cite{reimers2019sentence} MPNET~\cite{song2020mpnet} for embedding~\cite{silcock2022noise}, E5-Large~\cite{wang2022text} for embedding~\cite{kaddour2023minipile}, and OPT-125M~\cite{zhang2022opt} for embedding~\cite{abbas2023semdedup,tirumala2023d4}. 
These semantic deduplication processes are usually after the exact deduplication and approximate deduplication processes, removing the semantic duplicates by clustering the embedding points and keeping representative data in each clusters.
However, how to choose an appropriate pre-trained foundation models and whether this models can embed semantic information efficiently are still open questions. 

For image-text pairs or interleaved image-text documents, common methods of image deduplication include using the image URL~\cite{zhu2024multimodal} or employing pHash~\cite{zauner2010implementation} algorithms. It's worth noting that in the realm of interleaved image-text documents, there is currently no universally accepted, reliable method for deduplication based on both image and text elements simultaneously.

For Video multimodal data, video fingerprinting technology is uniquely applied to videos to identify and remove exact duplicates, addressing the challenge of video reuploads and mirrored content, which is more prevalent and complex in videos than in static images\cite{xu2023youku}.

\subsubsection{Data Enhancement}
Data enhancement of multimodal datasets usually focus on two aspects, enhancing the X modality data, and enhancing the text data. Traditional data augmentation methods for single modality data have been discussed thoroughly in previous work~\cite{xu2023comprehensive,cauli2022survey,ko2015audio,wei2020comparison,zhao2021optimization,feng2021survey}. In this work, we only consider data enhancement methods for MLLMs datasets. 

For vision-language models, enhancing the quality of image-caption datasets is crucial for training MLLMs. Improving the captions not only enhances their alignment with the images but also prevents the discarding of highly informative images due to poor text quality. Rewriting text using BLIP2 has been shown to effectively improve training results~\cite{Nguyen2023ImprovingMD}. However, for large-scale image-text datasets (approximately 1.28 billion), this method experiences diminishing returns in terms of ImageNet accuracy, possibly due to the lack of diversity in the generated captions~\cite{Nguyen2023ImprovingMD}. Approaches like LaCLIP~\cite{fan2024improving} and VeCLIP~\cite{Lai2023VeCLIPIC} employ LLMs with carefully designed prompts to generate more diverse captions.  ShareGPT4V~\cite{chen2023sharegpt4v} employed the 100K high-quality captions generated by GPT4-Vision to fine-tune an alternative caption model and named it as Share-Captioner. The Share-Captioner is capable of generating highly content-related descriptions with unified instruction for the pre-train dataset. MLM~\cite{wang2024finetuned} used GPT-4 or GPT-4V to constructing multimodal instruction tuning data on proposed quality scoring tasks to fine-tune MLM to realize accurate quality assessment. Then they adopt the fine-tuned MLM Filter to generate quality scores for each data point in the data pool and then select the high-quality data.

Improving image resolution can enhance the performance of MLLMs. Initially, MLLMs primarily processed fixed, lower-resolution inputs, typically around 224 pixels \cite{liu2024visual, chen2023shikra, zhu2023minigpt4}. Recent models such as LLaVA-1.5 \cite{liu2023improved} and BLiVA \cite{hu2024bliva} have improved performance by increasing the input resolution to 336 pixels and integrating task-specific global features. Furthermore, models like Qwen-VL \cite{bai2023qwen} and OtterHD \cite{li2023otterhd} have pushed resolution support to 448 pixels, incorporating fine-tuning in the visual encoder during training, while maintaining the original image size during inference, leading to more precise segmentation recognition. Notably, Monkey \cite{li2023monkey} has significantly raised the resolution to 896 pixels by employing multiple visual encoders and leveraging fine-tuning techniques from Qwen-VL. Models like LLAVA HR \cite{luo2024feast} and Vary \cite{wei2023vary} have introduced additional visual encoders to capture more complex features, requiring extensive pre-training tasks. LLAVA UHD \cite{xu2024llava} and Ureader \cite{ye2023ureader} have improved the model's capacity to understand detailed image features through adaptive region segmentation. However, a fundamental challenge remains within the MLLM architecture: models using lower-resolution inputs struggle to detect fine details, whereas those with higher resolutions may underperform in tasks requiring a broader global understanding.


\subsection{Commonly-used Datasets for MLLMs}
In this section we will briefly introduce commonly used multimodal datasets from different modality. These datasets are not utilized for this research.
The comprehensive summary of text-only datasets has been elaborately discussed in previous literature~\cite{liu2024datasets}; therefore, this part will not be extensively covered within the scope of this discussion. 

\subsubsection{Image Datasets}
For vision langauge models, the commonly used image-text datasets can be categorized into three distinct types: image-caption datasets, interleaved image-text datasets, and visual question answering datasets.
\paragraph{Image-Caption Datasets.}
Image-caption pairs datasets are the most commonly used datasets in the pre-training stage of MLLMs. By leveraging the captions of each image, one can align the image modality with the text modality, enabling LLMs to understand image information. 

General image-caption datasets contain images with short captions, typically one sentence or a few words, that describe the key features of the image. 
Notably, the LAION series (comprising LAION-400M~\cite{schuhmann2021laion} and LAION-5B~\cite{schuhmann2022laion}) represents some of the largest compilations, aggregating billions of pairs from the CommonCrawl. These datasets emphasize the maintenance of high quality through meticulous filtering based on the relevance of image-text pairs and the exclusion of inappropriate content. Derived from the LAION-5B subset, the LAION-COCO~\cite{schuhmann2022laion} dataset is specifically designed to explore the impact of synthetic captions on model training. 
Also curated from CommonCrawl project, the COYO-700M~\cite{kakaobrain2022coyo-700m} dataset emphasizes the informative correlation between images and alt-texts. 
The DataComp~\cite{gadre2024datacomp} challenge introduces COMMONPOOL, a substantial multimodal dataset constructed from 12.8 billion image-text pairs that encourages innovation in dataset design. 
Apart from CommonCrawl project, some self-crawled datasets also play a vital role.
Several datasets are curated from images with alt attributes from webpages. For example, Conceptual Captions (CC3M~\cite{sharma2018conceptual} and CC12M~\cite{changpinyo2021conceptual}) use generalized alt-text from the web and are processed with multiple filters, containing 3.3 million and 12 million image-text pairs respectively.
Also leveraging alt-text for images on webpages, ALT200M~\cite{hu2022scaling} dataset collects 200 million images with their alt attributes, aiming for understanding scaling of vision-language models.
For multilingual image-caption datasets, a representative one is the Wukong~\cite{gu2022wukong} dataset, which addresses the scarcity of large-scale datasets in the Chinese language, offering 100 million quality-assured pairs that are instrumental for the development of Chinese vision-language pre-training models. 
Larger dataset such as Long text \& image pairs (LTIP)~\cite{alayrac2022flamingo} dataset provides a unique collection of 312 million images with lengthier textual descriptions, enhancing models like Flamingo that are tailored for complex multimodal tasks.

Content descriptive image-caption datasets contain longer caption with more descriptive information. Usually each image in content description image-caption datasets contain at least 5 sentences of captions.
Content descriptive image-caption datasets are essential for advancing the interface between visual content and text descriptions, enhancing the ability of MLLMs to interpret and generate more detailed description from images. 
For example, MS-COCO~\cite{chen2015microsoft} employs images with multiple reference captions to refine evaluation metrics, while Flickr30K~\cite{young2014image} extends the variety of descriptions to bolster semantic inference capabilities. 
In addition, datasets like Visual Genome~\cite{krishna2017visual} provide comprehensive annotations beyond simple captions, including objects, attributes, and relationships, to facilitate complex scene understanding. While the AI Challenger Captions~\cite{wu2017ai} dataset specifically addresses the need for non-English language representation in image captioning, offering extensive annotations in Chinese to bridge the semantic gap between low-level visual features and high-level conceptual descriptions.
Narrative and textual comprehension are further introduced by datasets like VIST~\cite{huang2016visual}, which emphasizes storytelling through visual sequences. TextCaps~\cite{sidorov2020textcaps} incorporates reading comprehension into captioning. Collectively, these datasets not only propel advancements in automated image captioning but also tackle broader challenges in AI's capability to process and generate meaningful visual and textual data, significantly enhancing context-aware machine understanding.

\paragraph{Interleaved Image-Text Documents.}
A fundamental contrast between image-text caption datasets and interleaved image-text document datasets lies in their composition: image-caption pair datasets typically consist of a single image accompanied by multiple closely related captions, whereas interleaved image-text datasets consist of a text document interspersed with several illustrative images. The correlation between the images and the accompanying text tends to be relatively lower, but these datasets usually provides more semantic information, which is crucial for maintaining models' semantic understanding.
M3W~\cite{alayrac2022flamingo} leverages data from 43 million web pages, integrating images with text to train models in few-shot learning scenarios. MMC4~\cite{zhu2024multimodal}, an extension of the C4 corpus, aligns images with texts using CLIP features, resulting in a dataset containing over 101.2 million documents. OBELICS~\cite{laurenccon2024obelics}, sourced from CommonCrawl, comprises 141 million web pages.
OmniCorpus~\cite{li2024omnicorpus} scales the interleaved image-text datasets to 10 billion-level, providing much richer image and text information from diverse sources.

\paragraph{Visual Question Answer (VQA) Datasets}
Visual question answer (VQA) datasets facilitate advanced research in MLLMs by allowing models to interpret images and answer related questions. These datasets typically consist of images paired with corresponding questions and answers, aiming to develop and evaluate models' ability to understand visual content and provide accurate responses. 
For example, VQAv2.0~\cite{goyal2019making} expands on its forerunner by introducing complementary images that prompt different answers to the same question, effectively addressing biases and increasing the dataset's diversity. Visual-7W~\cite{zhu2016visual7w} extends the scope by incorporating questions across multiple dimensions—what, where, when, who, why, how, and which—linked to specific objects within images. ST-VQA~\cite{biten2019scene} integrates scene text into the VQA framework, making it possible to answer text-based questions from visual data. Shikra-RD~\cite{chen2023shikra} leverages advanced language models to annotate images with relational descriptions, enhancing image comprehension. OCR-VQA focuses on reading text from book covers to answer related questions, combining elements of optical character recognition and VQA. DocVQA~\cite{mathew2021docvqa} targets document images for extractive question answering, emphasizing precision in answers derived from visible text. A-OKVQA~\cite{marino2019ok} introduces questions that demand commonsense and world knowledge, pushing the envelope on reasoning capabilities required from AI systems. TextVQA~\cite{singh2019towards} specifically challenges models to read and understand text within images to respond accurately, and GQA~\cite{hudson2019gqa} promotes advanced reasoning over detailed visual scenes annotated with comprehensive scene graphs. These datasets collectively advance the field by enabling more nuanced interactions between AI models and the rich content within images, aiming for greater depth in the understanding and contextual integration of visual and textual data.

\subsubsection{Video Datasets}
Video datasets are comprehensive collections of video clips accompanied by associated annotations or labels. These datasets are meticulously designed to facilitate the training and evaluation of models for a wide range of video-related tasks, including action recognition, various video-text tasks, as well as video-centric dialogue.

Commonly used video-text dataset such as the MSR-VTT~\cite{bain2021frozen} features 10,000 video clips totaling 41.2 hours, each annotated with multiple descriptive sentences. Similarly, the WebVid series offers millions of web-sourced video clips with accompanying captions, expanding from 2.5 million pairs in WebVid-2M to 10 million pairs in WebVid-10M~\cite{bain2021frozen}, covering a vast array of 13,000 hours of video content. The VTP~\cite{alayrac2022flamingo} dataset further enriches the field with 27 million short video-text pairs sourced from a select few high-quality websites. Moreover, the recently introduced Panda-70M~\cite{chen2024panda} dataset contains a massive 70 million high-quality video-caption pairs, designed to enhance the training of high-performance MLLMs. Collectively, these datasets are integral to advancing the capabilities of MLLMs in understanding and interacting with video content. InternVid~\cite{wang2023internvid} scales video-text dataset to 230M annotated video-text pairs, with every video lasting 351.9s on average.

\subsubsection{Audio Datasets}
Audio datasets are curated collections of sound recordings along with associated annotations or labels. These datasets are specifically designed to support the training and evaluation of models for various audio-related tasks, such as speech recognition, sound event detection, music classification, and speaker identification. Audio datasets vary significantly in size, quality, and scope, with some focusing on specific languages or acoustic environments, while others strive for diversity to ensure generalizability across different audio processing scenarios. Notable examples include AISHELL-2~\cite{du2018aishell}, a Mandarin Chinese speech corpus with over 1000 hours of data from multiple regions; WavCaps~\cite{mei2023wavcaps}, an extensive English audio captioning dataset featuring approximately 400,000 clips; and VSDial-CN~\cite{chen2023x}, a multi-modal dataset derived from VisDial, encompassing visual data alongside related dialogues and captions, tailored for Automatic Speech Recognition (ASR) systems.

\subsubsection{3D Datasets}
The development of MLLMs benefits from the utilization of diverse 3D datasets, which provide comprehensive environmental and structural data critical for tasks like scene understanding and semantic segmentation. Notably, the ScanNet~\cite{dai2017scannet} dataset offers an extensive collection of RGB-D video data across 1,513 indoor scenes, annotated with 3D camera poses, surface reconstructions, and semantic segmentations, totaling over 2.5 million views. Similarly, the S3DIS~\cite{armeni2017joint} dataset includes point clouds from six large indoor areas, encompassing 271 rooms, with detailed semantic annotations for each point. The Structured3D~\cite{Structured3D} dataset provides a vast repository of 3,500 home designs, encompassing 21,835 rooms detailed with elements such as object geometry, materials, and textures, tailored for analyses in 3D reconstruction and interior design. 


%% file: section/sec4_pretrain.tex
\section{Data-centric pre-training}\label{sec:pre}
The pre-training stage of multimodal large language models (MLLMs) is crucial for developing models that can effectively process and generate information across multiple modalities. This stage can be divided into two distinct phases, each focusing on specific aspects of the model's architecture and training objectives.

The first phase involves two parallel processes: pre-training the LLM backbone (such as Vicuna~\cite{vicuna2023} and LLaMA2~\cite{touvron2023llama}) using text-only datasets, and pre-training the modality encoders (such as ViT~\cite{dosovitskiy2020image} and CLIP-ViT~\cite{radford2021learning} for visual encoding, C-Former~\cite{chen2023x} and HuBERT~\cite{hsu2021hubert} for audio encoding, and ULIP-2~\cite{xue2022ulip,xue2023ulip2} for 3D point cloud encoding) using pairs of data from different modalities, such as image-text or video-text pairs. This phase aims to establish a strong foundation for the model's understanding of both textual and non-textual information.

The second phase builds upon the knowledge acquired in the first phase by further training the LLM and modality encoders using a mixture of multimodal data. During this phase, the input projector (such as a linear projector, cross-attention mechanism, Q-Former~\cite{li2022blip}, or P-Former~\cite{jian2024bootstrapping}) is trained to effectively map the features extracted by the modality encoders into the LLM's embedding space. This mapping allows for a unified representation of textual and non-textual information within the model's architecture. Some researchers consider employing a selective training approach during the second phase, which involves keeping certain components of the model frozen while training specific parts. This approach helps to preserve the knowledge acquired during the first phase, allows for targeted adaptation to specific multimodal tasks, and reduces computational costs.

\subsection{Domain Mixture}

The performance of a language model (LM) is significantly influenced by the composition of its pre-training data, which often includes sources such as Wikipedia, books, and web text~\cite{xie2024doremi,xie2024data}.

Previous methods typically select domain weights heuristically or optimize them using downstream tasks~\cite{du2022glam}. However, these approaches can be suboptimal or costly, requiring the training of LMs for different sets of domain weights and potentially leading to overfitting to specific downstream tasks. DoReMi~\cite{xie2024doremi} offers a solution by optimizing domain weights using proxy models that are 30 times smaller than the target LLM, without requiring knowledge of downstream tasks. Recent work~\cite{liu2024regmix} formulate domain mixture problem as a regression problem. They assume the rank invariance of domain mixture between small models and large models, thus developing domain mixture regression function to predict the optimal domain mixture for model training. There are also some work discuss this problem through domain mixture scaling law. These work extend the original scaling law~\cite{kaplan2020scaling,hoffmann2022training} to domain data mixing law~\cite{que2024d, ye2024data, ge2024data}, discussing the relationship between domain mixture rate and loss through scaling models.

In the realm of video understanding, there is a consensus on the necessity for diverse datasets~\cite{xu2023youku,chen2024vast, song2023moviechat, chen2023videollm,wang2023internvid}. \citet{xu2023youku} highlight the use of hierarchical multi-label classification models~\cite{giunchiglia2020coherent} to ensure dataset balance and comprehensiveness across multiple dimensions, thereby optimizing the model's ability to understand and process various video domains, which is instrumental in training more versatile applicable video-language models.

\subsection{Modality Mixture}
When pre-training MLLMs, determining the optimal proportions of multimodal data is crucial for enhancing the model's performance across different tasks. MM1~\cite{mckinzie2024mm1} explores the balance between image-caption pairs, interleaved image-caption documents, and text-only data for vision-language pre-training. They found that a ratio of 5:5:1 for caption/interleaved/text data yields the best overall performance on text-only tasks as well as zero-shot and few-shot image-text tasks. Additionally, incorporating synthetic data~\cite{lai2023scarcity} improves the model's few-shot learning capabilities for image-text tasks.

For the development of MLLMs tailored for video understanding, the strategic integration of both video-text and image-text pairs during the pre-training phase is pivotal. This approach, highlighted in recent research~\cite{lin2023video,zhang2023video, jin2023chat, han2023autoad,li2023videochat,wang2023internvid}, enhances the model's capabilities by leveraging the complementary nature of these data types. Integrating labeled image datasets like COCO with video data broadens the variety of training datasets and addresses the scarcity of high-quality video-text resources. Studies~\cite{lin2023video, jin2023chat} indicate that training on both images and videos enhances models' ability to understand static and dynamic visual information without needing task-specific adaptations. Adding a temporal modeling module to the vision encoder bridges the temporal dynamics of videos with the static nature of images, fostering cohesive visual understanding across different types of visual media. However, treating images as single-frame videos~\cite{zhang2023video,luo2023valley} or pseudo-videos may weaken the model's ability to grasp the temporal aspects of video sequences. An imbalanced mix could also bias the model towards one modality, complicating its learning process. Therefore, determining the optimal mix ratio of image-text pairs and video-text pairs is crucial for improving the model's comprehensive understanding capabilities, ensuring a balanced approach to learning from both static images and dynamic videos. Additionally, video LLMs are increasingly being trained through multiple branches~\cite{chen2024vast,zhang2023video,lyu2023macaw}, such as vision-language, audio-language, and subtitle-language. These methods leverage the correlations among different modalities to broaden the models' understanding and learning capabilities. Recent studies~\cite{shu2023audio,chen2024vast,han2023autoad} have shown that combining audio, subtitles, and visual data during training enhances performance across various video understanding benchmarks. This modular training approach also offers the flexibility to pre-train with partial data~\cite{han2023autoad, sun2023fine}, such as visual-only or audio-only, particularly when comprehensive multimodal data is not available.

\subsection{Quality Selection}
Due to the diverse distribution of data, training a large model with all available data is not always optimal. Therefore, data selection becomes essential, offering benefits such as reduced training time and energy consumption~\cite{gadre2024datacomp}. Previous work has studied data selection methods using n-gram similarity with high quality datasets~\cite{xie2023data,gao2020pile,chowdhery2023palm}, perplexity~\cite{marion2023less, wenzek2020ccnet}, influence functions~\cite{park2023trak} and llm-based classifier~\cite{wettig2024qurating}. Unlike pure text datasets, data selection for multimodal datasets must consider the alignment between different modalities.
Data selection methods can be categorized into two types: active learning-based and pre-training selection. Active learning-based methods, such as CiT~\cite{Xu2023CiTCI}, use data proxies to dynamically select training data during the training process, achieving significant acceleration effects. Pre-training selection methods, like Datacomp~\cite{gadre2024datacomp} and Bunny~\cite{he2024efficient}, evaluate and select all data before training begins.

Regarding data selection criteria, methods can be distribution-agnostic, focusing solely on individual data point quality, or distribution-aware, considering the overall data distribution. Distribution-agnostic methods include training a model with the top 30\% of data ranked by CLIP score, which can significantly improve results~\cite{gadre2024datacomp}. More comprehensive metrics have also been developed, showing better performance than using CLIP score alone~\cite{wang2024finetuned}. For instance, \citet{mahmoud2024sieve} use the difference between original and synthetic captions generated by a small model to evaluate image-text pair alignment. Given the limited scope of purely distribution-agnostic work, combining distribution-agnostic and distribution-aware methods is often more effective. For example, using CLIP-score-based filters alongside image-based filters can outperform either method alone on large datasets~\cite{wang2024variance,gadre2024datacomp}.

\FloatBarrier

%% file: section/sec5_adaptation.tex
\section{data-centric adaptation}\label{sec:adap}
Adaptation is crucial for aligning pre-trained multimodal large language models (MLLMs) with specific tasks and user preferences. Self-supervised pre-training provides LLMs with a broad understanding of textual and multimodal information, while supervised fine-tuning (SFT) and reinforcement learning from human feedback (RLHF) are essential for adapting these models to excel in targeted applications and adhere to user preferences and societal values. Data-centric SFT involves training models on carefully curated datasets across multiple modalities. In contrast, data-centric RLHF focuses on collecting human judgments to guide the model toward generating desirable and ethically aligned responses. Emphasizing data-centric approaches highlights the importance of high-quality, domain-specific, and multimodal datasets in the successful adaptation of MLLMs.

\subsection{Data-Centric Supervised Finetuning}
Supervised fine-tuning (SFT) has become an indispensable technique for adapting MLLMs to specific domains and tasks. By leveraging carefully curated instruction-formatted datasets that encompass multimodal information, these models can be guided to acquire the necessary knowledge and abilities to excel in targeted applications.

During the SFT stage, various strategies have been explored, including fine-tuning one or all of the three primary pre-trained components: the LLM backbone, the modality encoders, and the input projector. This selective approach allows for a balance between adapting the model to specific tasks and preserving the generalized knowledge acquired during pre-training.

To facilitate effective supervised fine-tuning, diverse datasets that align with the target domain and task are used. These datasets typically include a combination of multimodal instruction-response pairs and text-only SFT data. By exposing the MLLMs to this rich variety of data, the model can develop a comprehensive understanding of the target domain and acquire the necessary skills to perform the desired tasks effectively.

The curation of self-supervised fine-tuning datasets usually includes four steps:
\begin{itemize}
    \item Collect X-text pairs from various data sources, similar to the pre-training stage as introduced in Section~\ref{pre:collect}.
    \item Process the data for further use, including filtering, deduplication, translation, etc.
    \item Construct the processed X-text pairs into instruction-response form, which involves designing the instruction and converting captions into answers.
    \item Select high-quality instruction-response pairs based on the fine-tuning needs. This step is not always necessary but can enhance the dataset's quality.
\end{itemize}

\subsubsection{SFT Data Collection and Processing}
\label{sec:adap:sft:datacol}

After collecting these data, several general approaches are employed to improve data quality, including filtering, deduplication, and data enhancement. These steps are similar to the data processing methods used in the pre-training stage described in Section~\ref{sec:pre:datapro}.

In addition to these general data processing approaches, specific methods aim to improve data quality for supervised fine-tuning tasks. The main goal is to add more textual information to the datasets or improve the quality of captions. These methods, while akin to the data enhancement techniques mentioned in Section~\ref{sec:pre:datapro}, are more focused on SFT data generation.
For example, MiniGPT-4~\cite{zhu2023minigpt4} curated detailed image description datasets for vision-language adaptation by leveraging their pre-trained model with instructions such as "Describe this image in detail," ensuring generated sentences contained more than 80 tokens. They also used ChatGPT to refine the descriptions and manually checked for quality, obtaining 3,500 high-quality image-text pairs.
Similarly, ShareGPT4V~\cite{chen2023sharegpt4v} used GPT-4 Vision to generate 100,000 high-quality captions that included world knowledge, spatial information, aesthetic evaluation, and more. These captions were directly used to generate SFT instruction-response pairs.

\subsubsection{Instruction-Response Pairs Generation}
After collecting and processing the original X-Text datasets, the next step is to generate the instruction-response pairs datasets based on the original datasets.
The supervised fine-tuning stage is typically aiming to enhance the model's ability on downstream tasks. 
For this reason, previous work consider to generate instruction-response pairs for different downstream tasks, including descriptive tasks, question-answer tasks, reasoning tasks, and classification tasks. Based on task categorization, previous work selects different source datasets and uses various methods to generate the instruction-response pairs. 
In this section, we first introduce the differences between each downstream task, and then we summarize representative methods of generating instruction-response datasets for each type of downstream task.

\paragraph{Captioning Instruction-Response Datasets.}
The simplest tasks are the captioning ones. For this type of task, images (or other modality data) are usually provided, and the MLLMs are asked to give out description of the data. 
The instruction-response datasets designed to enhance the model's captioning ability are relatively easy to construct. These kinds of instructions are usually one question asking for the description of the X modality data. 
Specifically, for image data, MiniGPT-4~\cite{zhu2023minigpt4} construct description instruction tuning by randomly sample instructions such as "Describe this image in detail", and their modified high-quality captions described in Section.~\ref{sec:adap:sft:datacol} are directly used as the response. 
While LLaVA~\cite{xu2024llava} considers designing two types of description instructions, brief description and detailed description and uses GPT-4~\cite{openai2023gpt} to generate the answers. For brief descriptions, they add keywords such as "concisely", "brief", and "short" in their questions, and for detailed descriptions, they add keywords such as "descriptive" and "comprehensive". 
ShareGPT4V~\cite{chen2023sharegpt4v} replaces the detailed description data in LLaVA with its GPT4-Vision generated high-quality captions.
X-InstructBLIP~\cite{panagopoulou2023x} further uses these descriptive prompts not only on image-text datasets but also on audio-text, video-text, and 3D-text datasets as a description instruction.

\paragraph{Question Answer Instruction-Response Datasets.}
Question answer task requires the model to answer questions based on the given information on the X modality. This kind of task includes simple questions answers, and multiple choices.
The Most well-known question-answer task is the visual question answer (VQA) task for visual modality. 
For VQA tasks, previous research on computer vision has developed a large amount of VQA datasets, and recent work on MLLMs leverages these VQA datasets for constructing instruction-response datasets. 
For example, LLaVA-1.5~\cite{liu2023improved} leverage GQA~\cite{hudson2019gqa}, OCR-VQA~\cite{mishra2019ocr}, OKVQA~\cite{marino2019ok} and VQAv2~\cite{goyal2019making} to enrich their instruction response datasets. They format these datasets in a uniform template as "Answer the question using a single word or phrase." TextVQA~\cite{singh2019towards} is another source used for constructing instruction-response datasets which contain text information in images. M3IT~\cite{li2023m} utilize VQA datasets, such as VQA-v2~\cite{goyal2019making}, Shapes VQA~\cite{andreas2016neural} and DocVQA~\cite{mathew2021docvqa}, to develop a multimodal instruction tuning dataset that contains 2.4 million instances. In addition to visual question-answer tasks, instruction tuning to improve model performance in the video question answer task also leverages previous video question answer datasets, such as MSRVTT-QA~\cite{xu2017video}, iVQA~\cite{yang2021just}, MSVD-QA~\cite{xu2017video}, and ActivityNet-QA~\cite{yu2019activitynet}. 

\paragraph{Reasoning Instruction-Response Datasets}
Shikra-RD~\cite{chen2023shikra} focuses on enhancing the models' ability for referential dialogue problems, i.e., users point to specific areas and ask questions. They collect public VQA datasets, image captioning data, and several datasets with existing positional annotations to build instruction tuning datasets. Leveraging captions from the Flickr30K~\cite{young2014image} image-caption dataset and using GPT-4, they obtain high-quality RD annotations.
PVIT~\cite{chen2023position} converts the traditional VQA datasets such as GQA and OCR into instruction-response datasets with region-level information.
MiniGPT-V2~\cite{chen2023minigptv2} designs a multi-task instruction template, following the conversation template from LLaMA-2~\cite{li2023llama}. They add task identifier tokens and spatial location representations to identify the task and spatial information in the image. They build five types of instruction-response datasets, including LLaVA-Instruct-150K, grounded image captions from Flickr30K~\cite{young2014image} with direct instruction and object parsing, multi-round conversations, and text-only unnatural instruction datasets~\cite{honovich2022unnatural}.
CogVLM~\cite{wang2024cogvlm} provides an IT dataset called CogVLM-SFT-311K. To construct this dataset, they manually select high-quality IT data from MiniGPT-4 and integrate it with LLaVA-Instruct-150K, translating it into Chinese. They then manually correct any noise in their collected instruction tuning datasets and translate them back into English. In their model training, they also leverage VQA datasets and bounding box instruction tuning datasets generated from four types of datasets: grounded captioning (GC) datasets with box-noun phrase pairs in each image, referring expression generation (REG) datasets with box and explanation, referring expression comprehension (REC) datasets with boxes annotated for text, and grounded visual question answering (GroundedVQA) datasets.
LLaVA1.5~\cite{liu2023improved} leverage region level VQA datasets such as Visual Genome~\cite{krishna2017visual} and RefCOCO~\cite{kazemzadeh2014referitgame, mao2016generation} in their IT datasets construction.

\paragraph{Other Instruction-Response Datasets}
Another type of downstream task is the classification task, where the data is classified into either a label from a candidate pool or an open-label.
M$^3$IT~\cite{li2023m} leverages various datasets for image classification, including ImageNet~\cite{russakovsky2015imagenet}, Grounded Object Identification (COCO-GOI)\cite{lin2014microsoft}, COCO-Text\cite{veit2016coco}, Image Text Matching (COCO-ITM)\cite{lin2014microsoft}, e-SNLI-VE\cite{kayser2021vil}, Multi-modal Fact Checking (Mocheg)\cite{yao2023end}, and IQA\cite{duanmu2021quantifying}.
Additionally, for image classification tasks, InstructBLIP~\cite{dai2024instructblip} uses HatefulMemes~\cite{kiela2020hateful}, a binary classification dataset for detecting hateful content in memes, to generate an instruction-response dataset.
X-InstructBLIP~\cite{panagopoulou2023x} further extends this approach to audio classification by using AudioSet~\cite{gemmeke2017audio}. The instructions used are sentences asking for the classification of the audio, such as "Classify the following audio". And the response is the class of the original data.

\subsubsection{SFT Data Selection}
Supervised fine-tuning is crucial for adapting LLMs with downstream applications. \citet{zhou2024lima} found that the primary knowledge of LLMs is acquired during the pre-training stage, and the purpose of instruction tuning is to enable LLMs to learn how to perform well on specific tasks and interact with humans. They determined that only a small set of carefully crafted, high-quality instructions is sufficient to endow LLMs with powerful instruction-following capabilities. Subsequently, various methods have been proposed for data selection to identify high-quality data, improving performance and reducing training costs. These methods are categorized into four types: coreset-based, LLMs-based, gradient-based, and self-instruction-based methods.

\paragraph{Coreset-Based Methods}
Coreset provides a compact representation of a larger dataset while preserving its essential characteristics. From a geometric perspective, similar data points in the feature space are close to each other. For example, \citet{sener2017active} employs the greedy k-center algorithm to select a coreset and applies it to CNN image classification.
In addressing the multi-class classification task, MODERATE CORESET~\cite{xia2022moderate} scores data points based on their distance to the class center. Data points with scores close to the score median are selected as the coreset, which achieves a balance between discriminative, compressible, and diverse. SIMILAR~\cite{kothawade2021similar} picks data points according to the submodular information measures (SMI). A properly selected SMI function can maintain the diversity of selected data and handle imbalance classes, out-of-distribution data, and redundancy. Some algorithms select data points based on model performance, focusing on "important" samples.
For example, \citet{bachem2017practical} perform importance sampling based on the upper bound of the sensitivity score to generate a coreset. Generally speaking, data points with high costs are more likely to be selected.
Other mertics of performance include GraNd~\cite{paul2021deep}, least confidence~\cite{wang2014new}, and etc.

Given that Coreset techniques have demonstrated promising results both theoretically and empirically, it's feasible to apply these algorithms to the data used for training LLMs. \citet{chen2023maybe} employ the K-greedy algorithm to identify just 0.5\% of the core data for fine-tuning a pre-trained language model, achieving performance only 1-2\% lower than that obtained using the entire dataset. Similarly, the approach proposed by~\citet{das2023deft} involves selecting data that represents the easiest and most challenging examples, a strategy proven effective in~\cite{sorscher2022beyond}. By utilizing only 32.5\% of carefully selected data, they achieved state-of-the-art results. 

\paragraph{LLMs-Based Methods}
External model-based methods frequently utilize external models to evaluate the quality, diversity, and complexity of data. These models serve various functions, from scoring data to enhancing its characteristics. For instance, \citet{du2023mods} leverage DeBERTa~\cite{he2020deberta} for scoring, retaining high-quality data, and combining it with the k-center greedy algorithm to select diverse data. \citet{chen2023alpagasus} score the accuracy of data using ChatGPT to pick out high-quality data. \citet{xu2023rethinking} use GPT-4 to rewrite data to increase their complexity and then streamline it by reducing its variety and improving its quality. \citet{liu2023makes} train two models using ChatGPT's labeled data to score the quality and complexity of the data. \citet{lu2023instag} rely on ChatGPT to tag each instance, defining its complexity and diversity based on these tags. \citet{parkar2024selectllm} first cluster the data, and then use GPT-4 to select high-quality data for each cluster. \citet{zheng2024pas} and \citet{sun2024efficient} leverage LLMs to automatically select high-quality domain data, achieving superior performance. For multimodal instruction tuning datasets, MLLMs are typically used to obtain or select high-quality instruction tuning data. LLaVA1.5~\cite{liu2023improved} pioneers the use of text-only GPT-4 to expand the COCO~\cite{lin2014microsoft} bounding box and caption dataset into a multimodal instruction-following dataset. Following LLaVA, ShareGPT-4v~\cite{chen2023sharegpt4v} utilizes GPT-4v to enhance a portion of LLaVA's instruction tuning data. For Video Large Language Models (VideoLLMs), \citet{liang2024keyvideollm} pioneered the use of CLIP score for video keyframe data selection, achieving SoTA performance.

\paragraph{Gradient-Based Methods}
Neural Networks are typically trained through gradient-based optimization techniques. Consequently, researchers have been exploring gradient-based methods to use a subset of data to approximate the gradient effectively. The EL2U score was introduced as a method to evaluate how removing a single data point affects the gradient~\cite{paul2021deep}. However, the EL2U score is limited to single-task learning scenarios. To address multi-task learning, particularly in NLP tasks, another approach has been developed~\cite{attendu2023nlu}. This method applies the EL2U score to each individual task and then combines these scores using the L2 norm to accommodate the complexities of multi-task learning. Another notable work is~\cite{xia2024less}, which adapts existing influence formulations to work with the Adam optimizer. It only uses a few steps of gradients to compute gradient features efficiently and then stores the compressed features in a gradient datastore for efficient data selection.

\paragraph{Self-Instruction-Based Methods}
For self-instruction-based methods, the evaluation and selection of data do not require the involvement of any external models. \citet{li2023quantity} proposes a self-instruction method for LLMs to autonomously identify and select challenging data. \citet{li2023one} identifies high-quality data as those that, when included as part of a few-shot example, can enhance the model's performance. \citet{kung2023active} proposes a novel task-level uncertainty metric that measures the sensitivity of LLMs to instruction perturbations for a task, to identify high-quality data, and combines this with the concept of active learning to iteratively select data. \citet{liu2024selectit} scores data at three granular levels: token, sentence, and model. Each level's scoring is calculated based on the previous level, with the model-level score serving as the final score.

\subsection{Data-Centric Human Preference Alignment}
To align human preferences with language models, previous work has considered constructing instruction-response datasets that respect human preferences. These datasets usually involve three main elements. RLHF aims to use reinforcement learning (RL) to align the model with human preferences~\cite{ouyang2022training}. After training the LLM, RLHF collects human feedback to rank Q\&A text data, trains a reward model (RM), and then uses RL to fine-tune the LLMs.

The most impactful work of RLHF is InstructGPT~\cite{ouyang2022training}. To collect human feedback, InstructGPT first designs prompts manually and from users who use the InstructGPT playgrounds. Then, they process the prompts using heuristic de-duplication methods, such as checking for prompts that share a long common prefix and limiting the number of prompts to 200 per user ID, to protect the user's sensitive information and improve the quality of prompts.

These prompts encompass multiple tasks such as text generation, question answering, dialogue, summarization, and more. Human taggers are then asked to rank the responses to each prompt, constructing a Q\&A dataset used to train the reward model (RM). Subsequently, the fine-tuning stage of the LLM is framed as a reinforcement learning (RL) problem using the trained reward model. In LLaMa2~\cite{li2023llama}, human preferences are further divided into fine-grained aspects, with two separate reward models trained to rank helpfulness and safety, respectively. Further work also aims to improve the models' generation abilities from these aspects.

For MLLMs, several works have focused on aligning models with human preferences. Based on LLaVA~\cite{liu2023visual}, LLaVA-RLHF~\cite{sun2023aligning} designed a reinforcement learning from human feedback (RLHF) method to align LLaVA with human feedback. DRESS~\cite{chen2023dress} aims to improve models' response quality from the perspective of human values, particularly the 3H criteria~\cite{ouyang2022training} (helpfulness, honesty, and harmlessness). They provide two datasets: the Large Vision Language Model with Natural Language Feedback (LVLM\_NLF) dataset and the Vision-Language Safety (VLSafe) dataset. To construct the VLSafe dataset, they adopt the LLM-Human-in-the-Loop approach, iteratively creating and filtering the data based on the COCO dataset. The LLM they used is GPT-3.5 Turbo. The final dataset contains 5,874 samples, meeting most requirements for harmlessness alignment and evaluation.

\FloatBarrier

%% file: section/sec6_evaluation.tex
\section{evaluation}\label{sec:eval}
To offer a comprehensive data-centric viewpoint for assessing MLLMs, this section begins with two key aspects. First, we review widely-used methods for evaluating data quality. Then, we provide an overview of MLLMs evaluation datasets, highlighting their collection, processing, and characteristics.

\subsection{Data Evaluation}\label{subsec:Data_Evaluation}
Evaluating datasets is essential to ensure their quality and reliability. This process helps identify and correct errors and biases, enhancing the accuracy of model training and predictions. In this section, we summarize recent dataset evaluation metrics from the following perspectives.

\subsubsection{Dataset Diversity}
Dataset diversity is crucial for developing robust and generalizable machine learning models. A diverse dataset enables the model to manage various scenarios and minimizes the risk of bias. Some studies~\cite{heusel2017gans, sajjadi2018assessing} require a reference distribution or a dataset. For instance,~\citet{heusel2017gans} measures the Wasserstein-2 distance between two Gaussian distributions: one fitted to the embeddings of the reference sample and the other to the embeddings of the sample being evaluated for diversity. Other work use similarity scores to define diversity. For example,~\citet{shen2019mixture, fomicheva2020unsupervised} utilize the average pairwise similarity score or its complement, the average dissimilarity.~\citet{fomicheva2020unsupervised} propose a two-metric evaluation paradigm using precision and recall, with precision measuring quality and recall assessing diversity in terms of coverage of the reference distribution.
The Vendi score~\cite{friedman2023vendi, pasarkar2023cousins, yeh2023navigating} stands out from existing diversity evaluation metrics as a reference-free, flexible, and interpretable metric that measures internal diversity without comparing it to a reference distribution. Its reliance on a user-defined similarity function allows for broad applicability across domains, incorporating correlations between features while being computationally efficient and unsupervised, which makes it a valuable tool for diverse machine learning applications. Oppositely, \citet{lee2023beyond} use diversity score called Task2Vec diversity coefficient~\cite{miranda2022curse} to evaluate the diversity of publicly available datasets.

\subsubsection{Dataset Quality}
Dataset quality is essential for the accuracy and reliability of machine learning models. High-quality data enables models to learn effectively and make precise predictions. 
Ensuring dataset quality is essential for the accuracy and reliability of machine learning models. High-quality data enable effective learning and precise predictions.
There are several approaches to evaluate data quality.
TRUE~\cite{honovich2022true} offers a thorough evaluation of factual consistency metrics in grounded text generation systems. By standardizing datasets and introducing a meta-evaluation protocol, it highlights the robust performance of large-scale NLI and QG-QA methods across different tasks.
Object Hallucination~\cite{rohrbach2018object} introduces CHAIR (Caption Hallucination Assessment with Image Relevance), which measures the proportion of words in a generated caption that correspond to objects present in the image, using ground truth sentences and object segmentations. 
FAITH SCORE~\cite{jing2023faithscore} assesses the faithfulness of generated answers from large vision-language models (LVLMs). It identifies descriptive statements, extracts atomic facts, and checks their consistency with input images. This structured process ensures that generated answers align closely with the visual content, providing a comprehensive evaluation of faithfulness in LVLM outputs.

\subsubsection{Dataset Similarity}
In machine learning theory, performance improves when the training data distribution closely matches the testing data distribution. Therefore, it is essential to evaluate the distances of data distribution both theoretically and empirically. Common metrics for measuring distribution similarity include Euclidean distance, KL division, CORAL loss~\cite{sun2016deep}, Wasserstein distance, and MMD distance~\cite{jiang2022transferability}.
The MAUVE scores~\cite{pillutla2023mauve} serve as a comparison measure between pairs of distributions, such as those encountered in generative modeling of text or images. This metric provides statistical bounds and extensive experiments to demonstrate its effectiveness.

\subsection{Evaluation Datasets for MLLMs}
In this section, we will briefly summarize different types of evaluation datasets according to downstream tasks.

Common evaluation tasks for MLLMs can be divided into captioning tasks, question answering tasks, perception and reasoning tasks, and other noteworthy tasks such as classification tasks.
Captioning datasets~\cite{chen2015microsoft,venugopalan2015translating, kim2019audiocaps} act as benchmarks to evaluate a model's fundamental ability to understand information from different modalities. They can be used to assess the zero-shot or few-shot learning capabilities of MLLMs in multimodality tasks, offering insights into their ability to generalize and adapt to new scenarios with limited training examples.
While evaluation datasets for question answering (QA) are crucial for assessing the performance of MLLMs that handle diverse data types like text, images, and audio~\cite{goyal2019making,xu2017video,lipping2022clotho}. These datasets require models to analyze and combine information from multiple sources to respond to relevant questions accurately.
Perception and reasoning are crucial tests for assessing the general abilities of machine learning language models (MLLM). These tasks requires model to reason for complicated logic and capture structured or spatial information~\cite{singh2019towards,masry2022chartqa,liu2023visual,mangalam2024egoschema}.
There are also several noteworthy tasks not mentioned above, including classification task that require model to classify examples for right labels~\cite{kiela2020hateful,piczak2015esc,piczak2015esc,wu20153d,goyal2017something}, and multimodal dialog task that require for model's in-context learning abilities to answer questions during multiple rounds of conversations~\cite{bai2023touchstone}.

%% file: section/sec7_future_direction.tex
\section{future direction}\label{sec:future}
In this section, our attention is directed towards research avenues for MLLMs from a data-centric viewpoint. We categorize forthcoming directions based on the structure of our article. We begin by addressing future directions concerning data collecting and processing, followed by pre-training, adaptation, and evaluation aspects.

\subsection{Data Processing System for MLLMs.}
The processing of data for MLLMs entails numerous intricate steps across various modalities, as outlined in Section \ref{sec:pre:datapro}. Recent efforts in preparing data for MLLM training involve processing data autonomously through specialized data pipelines and processing operators \cite{alayrac2022flamingo, changpinyo2021conceptual, bain2021frozen, du2018aishell}. While previous work has focused on designing data processing systems for LLMs, such as Data-Juicer \cite{chen2023data} and Oasis \cite{zhou2023oasis}, there remains a gap in the availability of data processing systems tailored specifically for MLLM data. Such systems have to be equipped to handle multi-modal data types, encompassing not only textual data but also images, videos, audio, and 3D data formats.

\subsection{Data Quantity Analysis for MLLM Pre-training.}
Large language models have shown specific properties such as 
Large language models exhibit distinct characteristics such as emergent abilities and scaling laws. Previous research has examined emergent phenomena concerning model scale, measured by factors like training compute and the number of model parameters, as well as in relation to evaluation metrics~\cite{wei2022emergent, schaeffer2024emergent}. Scaling laws have been investigated concerning model size, data scale, and data quality. OpenAI initially explored the power-law relationship of pre-trained loss with respect to model size, dataset size, and training compute in neural language models~\cite{kaplan2020scaling}. Subsequently, DeepMind introduced a new scaling law demonstrating the optimal allocation of compute resources~\cite{hoffmann2022training}. Building on this,~\citet{goyal2024scaling} investigated the trade-offs between dataset quality and quantity, while~\citet{ye2024data} examined scaling laws associated with data mixtures. Furthermore,~\citet{zhang2024scaling} analyzed the impact of these properties, particularly focusing on text-only LLMs. However, there remains a gap in understanding how data quantity influences emergent abilities in MLLMs. Exploring the scaling laws of MLLMs, particularly regarding the evaluation of data quantity, represents an area that warrants further investigation.

\subsection{Data Quality Analysis for MLLM Pre-training.}
Adjusting the ratio and quantity of pre-training data holds significant potential for enhancing model performance. However, with the abundance of pre-training data available, there arises a critical necessity for efficient data selection algorithms. Evaluating data for large models typically demands the utilization of the models themselves, which can impose significant computational burdens. Hence, the development of Proxy Models becomes crucial, as they offer substantial reductions in computational costs. Moreover, Proxy Models prove instrumental in optimizing the mixing ratio of data from diverse sources. While several approaches like Doremi~\cite{xie2024doremi} and Doge~\cite{fan2023doge} have been proposed in this regard, there remains a need for the refinement of proxy models and further exploration into the relationship between proxy models and their original counterparts.

\subsection{MLLM Data Evaluation.}
As elaborated in Section \ref{subsec:Data_Evaluation}, despite the introduction of various data evaluation metrics, there remains a notable absence of comprehensive metrics tailored specifically for evaluating multimodal data. The assessment of multimodal data presents heightened challenges owing to its diverse array of data types, often encompassing multiple tasks and modalities~\cite{liu2023improved, chen2024panda}. The quality of such data can be impacted by the individual models corresponding to each modality and their alignments. Hence, it becomes imperative to scrutinize the quality of multimodal data from multiple dimensions.
In addition to relying solely on human-defined features, enhancing data quality can be achieved by leveraging statistical metrics and methodologies, such as scrutinizing distributions. The notion of domain adaptation offers theoretical underpinnings for matching distributions~\cite{jiang2022transferability}. Consequently, assessing distributions emerges as a promising avenue, both in terms of theoretical frameworks and empirical validation.

\subsection{Data Quality Improving for MLLM Supervised Fine-Tuning.}
Harnessing Instruction Tuning data can substantially bolster a model's proficiency in adhering to instructions and enhancing its performance across specific tasks. Given the pivotal role of both quality and quantity in optimizing model performance, the exploration of this area emerges as a crucial research endeavor.
The evaluation of data for large-scale models necessitates metrics that are model-agnostic, effectively capturing the distinct characteristics of the data. As highlighted in the preceding paragraph, there is a pressing need to devise metrics tailored specifically for instruction tuning data.
An alternative pragmatic approach involves employing Large Language Models (LLMs) for data evaluation, leveraging the extensive knowledge accrued during their pre-training phase~\cite{du2023mods, chen2023alpagasus, xu2023rethinking, liu2023makes, lu2023instag, parkar2024selectllm}. However, utilizing LLMs for data assessment proves to be cost-effective. Despite the success of GPT-based methods for automated data quality evaluation, such approaches often lack interpretability. Hence, further investigation is warranted to gain a deeper understanding of GPT's efficacy and to delineate the boundaries of employing GPT for data evaluation.

\subsection{MLLM Lifelong Learning.}
MLLMs often learn from diverse data types sequentially across various training phases~\cite{li2022blip, liu2024visual, liu2023improved}. Typically, this sequential training starts with initializing the model using pre-trained weights for each modality, followed by pre-training, instruction tuning, and concluding with reinforcement learning from human feedback. It is critical throughout these stages to prevent catastrophic forgetting, ensuring the model retains its foundational language capabilities while acquiring multimodal skills. Therefore, integrating lifelong learning strategies with MLLMs represents a significant and necessary research direction.

%% file: section/sec8_conclusion.tex
\section{conclusion}\label{sec:conclu}
In this survey, we review recent advancements in data-centric multimodal large language models (MLLMs), introducing key concepts, findings, and techniques essential for processing training data for these models. Specifically, our discussion focuses on three crucial aspects of data handling: pre-training, adaptation, and evaluation. For each aspect, we highlight pivotal techniques and insights for data processing that are critical for effectively training MLLMs. Additionally, we provide a comprehensive summary of available data resources for training these models and discuss strategies for utilizing machine learning to enhance the performance of MLLMs. This survey aims to encapsulate the most recent literature on data-centric MLLMs, serving as a valuable reference for both researchers and engineers in the field.

%% file: section/appendix1.tex
\section{appendix}
\subsection{Evaluation datasets for MLLMs}\label{sec:appen:eval}
\subsubsection{Captioning Tasks}
Captioning datasets act as benchmarks to evaluate a model's fundamental ability to understand information from different modalities. They can be used to assess the zero-shot or few-shot learning capabilities of MLLMs in multimodality tasks, offering insights into their ability to generalize and adapt to new scenarios with limited training examples.

For image captioning tasks, numerous datasets are available to assess a model's capability in image understanding. MS-COCO~\cite{chen2015microsoft} is one of the most frequently used datasets for training and evaluating image captioning. Flickr30K~\cite{young2014image} is also extensively utilized to evaluate models' capabilities across Zero/Few/Full-shot scenarios.
The Karpathy split for MS-COCO and Flickr30K~\cite{karpathy2015deep} is commonly adopted to divide training and testing data in the training and evaluation of MLLMs.
Nocaps~\cite{agrawal2019nocaps} was developed to overcome the limitations of MS-COCO by focusing on objects not present in MS-COCO captions, serving as a valuable complement to the MS-COCO dataset.

For video captioning, the MSVD dataset~\cite{chen2011collecting} is a popular choice for evaluating action recognition and video description tasks~\cite{venugopalan2015translating}, comprising 2,089 video segments and 85,550 English descriptions, with a standard split provided by~\citet{venugopalan2015translating}. The MSRVTT dataset~\cite{xu2016msr} includes 10,000 web video clips totaling 41.2 hours, each annotated with approximately 20 natural sentences.
The DiDeMo dataset~\cite{anne2017localizing} contains over 10,000 unedited personal videos, each associated with 3-5 pairs of descriptions pinpointing specific moments. The VATEX dataset~\cite{wang2019vatex} includes more than 41,250 videos paired with 825,000 captions in English and Chinese, featuring over 206,000 English-Chinese parallel translation pairs. Additionally, the TVC dataset~\cite{lei2020tvr} includes both validation and private test components for further evaluation. These datasets are instrumental in evaluating both the video captioning and the video information retrieval capabilities.

For evaluating audio captioning ability, the two most commonly used datasets are AudioCaps~\cite{kim2019audiocaps} and Clotho~\cite{drossos2020clotho}. The AudioCaps dataset features 46,000 pairs of audio clips and human-written text descriptions. Meanwhile, the Clotho dataset includes 4,981 audio samples, each lasting between 15 to 30 seconds, accompanied by 24,905 captions that range from 8 to 20 words in length. Both datasets are instrumental in assessing a model's audio understanding capabilities.

\subsubsection{Question Answering Tasks}
Evaluation datasets for question answering (QA) are crucial for assessing the performance of MLLMs that handle diverse data types like text, images, and audio. These datasets require models to analyze and combine information from multiple sources to respond to relevant questions accurately.

Visual Question Answering (VQA) evaluation dataset tests the MLLM's ability to integrate and interpret both text and image data. These datasets contain images paired with questions that require visual understanding to answer. Commonly used VQA evaluation datasets such as
VQAv2~\cite{goyal2019making} contains 13 million answers associated with approximately 200,000 images from the COCO dataset. It also contains multi images question answer pairs that can test models' multi image understanding abilities.
GQA~\cite{hudson2019gqa} dataset is designed to address deficiencies in previous visual question answering (VQA) datasets by fostering advanced visual reasoning and reducing biases. It utilizes over 22 million questions generated from 113K real-world images annotated with detailed scene graphs, detailing objects, attributes, and relationships.
OKVQA~\cite{marino2019ok} was also created to address the limitations of existing VQA benchmarks by focusing on knowledge-based visual question answering. This dataset consists of more than 14,000 questions that require external knowledge to answer, covering various categories such as science \& technology, history, and sports.
Vizwiz~\cite{gurari2018vizwiz} focuses on collecting visual questions from blind users, presenting unique challenges such as image quality and conversational question styles.
POPE~\cite{li2023evaluating} focuses on evaluating object hallucination in vision-language models, using 6,136 binary questions to test if models hallucinate non-existent objects in images from MS-COCO and other sources.

Video question-answering evaluation datasets are developed to assess the capabilities of models in video-to-text translation and overall video comprehension. Key datasets include MSVD-QA~\cite{xu2017video} and MSRVTT-QA~\cite{xu2017video}, derived from the MSVD~\cite{chen2011collecting} and MSRVTT~\cite{xu2016msr} video captioning datasets, respectively, which are frequently utilized for video QA tasks. For more intricate web videos, the ActivityNet-QA~\cite{yu2019activitynet} dataset, containing 58,000 QA pairs from 5,800 complex videos, tests models' abilities to understand complex video content. The TGIF-QA~\cite{jang2017tgif} dataset, with 165,165 QA pairs from 71,741 animated GIFs, evaluates spatio-temporal reasoning and visual question-answering skills in videos. LSMDC~\cite{rohrbach2017movie}, which includes 118,114 sentences aligned with clips from 202 movies, is used for assessing video description generation and movie QA abilities. Additionally, MoVQA~\cite{rohrbach2017movie}, featuring 21,953 manually annotated QA pairs from 100 diverse movies, is aimed at evaluating the understanding of long-form videos over various temporal durations, focusing on complex and extended video content comprehension.

Audio question-answering evaluation datasets are designed to test audio-to-text translation and the ability to understand audio. The ClothoAQA~\cite{lipping2022clotho} dataset, derived from the Clotho dataset, includes 1,991 audio files, each lasting between 15 to 30 seconds, with six different questions per audio file collected. The MUSIC-AVQA~\cite{li2022learning} dataset contains more than 45,000 question-answer pairs extracted from more than 9,000 videos and more than 150 hours of content. It uses 33 question templates across 9 types of questions to support spatio-temporal reasoning in audio-visual scenarios.

\subsubsection{Perception and Reasoning Tasks}

Perception and reasoning are crucial tests for assessing the general abilities of machine learning language models (MLLM). TextVQA~\cite{singh2019towards} consists of 28,408 images with questions that require reading and reasoning about the text in the image. ChartQA~\cite{masry2022chartqa} offers various questions based on real-world charts, testing visual and logical reasoning. AI2D~\cite{kembhavi2016diagram} includes over 5,000 grade school science diagrams with extensive annotations to evaluate diagram interpretation skills. ScienceQA\cite{lu2022learn} presents around 21,000 multimodal questions from science curricula, emphasizing multi-hop reasoning. MathVista~\cite{lu2023mathvista} comprises 6,141 image-text pairs to benchmark mathematical reasoning in visual contexts. MMVet~\cite{yu2023mmvet} includes 200 images with questions that test large multimodal models in six capabilities. MMBench~\cite{liu2023mmbench} assesses 20 distinct abilities with 3,000 questions, ranging from object localization to social reasoning. LLaVAW~\cite{liu2024visual} features 24 images with diverse visual content and complex reasoning tests. MME~\cite{fu2024mme} focuses on 14 subtasks covering perception and cognition, such as text translation and arithmetic. Lastly, MVBench~\cite{li2023mvbench} contains 20 video tasks that challenge comprehension of spatial and temporal dynamics, designed to evaluate advanced understanding skills.

There are several datasets designed to evaluate MLLMs' spatial reasoning abilities. VSR~\cite{liu2023visual} comprises over 10,000 natural text-image pairs with 66 types of spatial relations, making it an excellent resource for evaluating spatial reasoning. The datasets RefCOCO, RefCOCO+, and RefCOCOg~\cite{kazemzadeh2014referitgame, mao2016generation} are crucial to assess how well models understand and ground natural language expressions within visual contexts, featuring images paired with expressions that describe specific objects or areas. These datasets, which vary in expression collection and annotations, challenge models to integrate vision and language by capturing detailed spatial relationships and contextual nuances. Additionally, the GRIT~\cite{gupta2022grit} benchmark addresses seven vision tasks using multiple data sources to enhance referring expression grounding. DocVQA~\cite{mathew2021docvqa} focuses on visual question answering within document images, utilizing a web-based tool to generate its 50,000 questions based on over 12,000 document images. OCR-VQA~\cite{mishra2019ocr} contains 207,572 images of book covers, with more than a million question-answer pairs, testing models' ability to understand text in document images and perform visual question-answering. Lastly, SEED-Bench~\cite{li2023seed} evaluates spatial and temporal comprehension in multimodal contexts, serving as a platform to test generative comprehension capabilities.

For video reasoning, there exist numerous evaluation datasets. EgoSchema~\cite{mangalam2024egoschema} features over 5,000 very long-form video language understanding questions derived from 250 hours of diverse egocentric video data. VideoChatGPT~\cite{maaz2023video}, which is used to tackle challenges in video-based conversation models, addresses aspects such as temporal understanding, spatial consistency, and contextual comprehension. Both EgoSchema~\cite{mangalam2024egoschema} and VideoChatGPT~\cite{maaz2023video} offer the means to assess models on longer and more intricate video content. Charades-STA~\cite{gao2017tall} comprises approximately 10,000 videos, each annotated with temporal activity details across 157 activity categories and multiple video-level descriptions. Additionally, QVHighlights~\cite{lei2021detecting} contains over 10,000 YouTube videos covering various topics such as everyday activities, travel, and social and political events, facilitating the evaluation of MLLMs' abilities in detecting moments and generating highlights.

\subsubsection{Other Noteworthy Tasks}
There are also several noteworthy tasks not mentioned above. One of the traditional tasks is the classification task. For the classification task, InstructBLIP uses HatefulMemes~\cite{kiela2020hateful}, a binary hateful content classification dataset for memes classification. ImageNet-1K~\cite{deng2009imagenet} is also a widely used dataset to test models' image classification ability. While X-InstructBLIP extends this to ESC50~\cite{piczak2015esc} for audio classification and ModelNet40~\cite{wu20153d} for 3D classification. Furthermore, M$^3$IT~\cite{li2023m} use Something-Something~\cite{goyal2017something} for video action classification.
Multimodal dialogue task is different from question answering task because it requires multiple rounds of conversion. TouchStone~\cite{bai2023touchstone} explores multimodal dialogue through detailed image annotations, addressing the challenge of evaluating LLMs in open-ended dialogues.

\subsection{Supervised fine-tuning datasets}
\begin{table}

    \centering
    \caption{A detailed list of datasets and processing methods for different models with image modality during fine-tuning stage. * indicates the dataset is newly generated using certain method within a respective model, while the other datasets (without *) serve as the original data sources for that model. - denotes directly using original data sources without additional processing. \#Examples denotes the statistics for each dataset.}
    \tiny
    \smaller
    \setstretch{0.92}
    \begin{tabular}{cclcl}
    \toprule
    
    \textbf{Models} & \textbf{Time} & \multicolumn{1}{c}{\textbf{Finetuning Datasets}} & \multicolumn{1}{c}{\textbf{Processing Method}} & \multicolumn{1}{c}{\textbf{\#Examples}} \\
    \midrule
    
    \multirow{3}{*}{Flamingo\cite{alayrac2022flamingo}}
    & \multirow{3}{*}{2022.4}
    & VQAv2 
    & \multirow{3}{*}{-} & \#images: 265K \hspace{0.5em} \#questions: 1.4M \\
    && VizWiz && \#image-question pairs: 33.8K \\
    && TextVQA && \#images: 28.4K \hspace{0.5em} \#questions: 45.3K \\
    \midrule
    
    \multirow{4}{*}{BLIP-2\cite{li2023blip}}
    & \multirow{4}{*}{2023.01}
    & COCO 
    & \multirow{4}{*}{-} & \#images: 330K \hspace{0.5em} \#captions: 5 per image \\
    && NoCaps && \#images: 15.1K \hspace{0.5em} \#captions: 11 per image \\
    && VQAv2 && \#images: 265K \hspace{0.5em} \#questions: 1.4M \\
    && Flickr30K && \#images: 31.7K \hspace{0.5em} \#captions: 5 per image\\
    \midrule
    
    \multirow{2}{*}{LLaVA\cite{liu2024visual}}
    & \multirow{2}{*}{2023.04}
    & COCO 
    & \multirow{2}{*}{GPT-4, Prompt, Manual examples} & \#images: 330K \hspace{0.5em} \#captions: 5 per image \\
    && LLaVA-Instruct-150K* && \#image-text pairs: 158K \\
    \midrule
    
    \multirow{2}{*}{MiniGPT-4\cite{zhu2023minigpt4}}
    & \multirow{2}{*}{2023.04}
    & CC12M 
    & \multirow{2}{3cm}{\centering Manual verifying and refining, ChatGPT, Random selecting} & \#image-text pairs: 12M \\
    && \text{cc\_sbu\_align}* && \#image-text pairs: 3.5K \\
    \midrule
    
    \multirow{1}{*}{mPLUG-Owl\cite{2023mplug-owl}}
    & \multirow{1}{*}{2023.04}
    & LLaVA-Instruct-150K 
    & \multirow{1}{*}{-} & \#image-text pairs: 158K \\
    \midrule
    
    \multirow{4}{*}{InstructBLIP\cite{dai2024instructblip}}
    & \multirow{4}{*}{2023.05}
    & ScienceQA 
    & \multirow{4}{*}{-} & \#image-question pairs: 10.3K \\
    && OCR-VQA && \#images: 207K \hspace{0.5em} \#image-question pairs: 1M \\
    && OKVQA && \#images: 14K \hspace{0.5em} \#questions: 14K \\
    && A-OKVQA && \#images: 23.6K \hspace{0.5em} \#questions: 24.9K\\
    \midrule
    
    \multirow{6}{*}{PaLI-X\cite{chen2023pali-x}}
    & \multirow{6}{*}{2023.05}
    & COCO 
    & \multirow{6}{*}{-} & \#images: 330K \hspace{0.5em} \#captions: 5 per image \\
    && NoCaps && \#images: 15.1K \hspace{0.5em} \#captions: 11 per image \\
    && TextCaps && \#images: 28.4K \hspace{0.5em} \#captions: 5 per image \\
    && VizWizCap && \#images: 39.1K \hspace{0.5em} \#captions: 5 per image\\
    && Screen2Words && \#images: 22.4K \hspace{0.5em} \#captions: 5 per image\\
    && WidgetCap && \#images: 21.7K \hspace{0.5em} \#captions: 162K \hspace{0.5em} \#widgets: 61.2K\\
    \midrule
    
    \multirow{2}{*}{Shikra\cite{chen2023shikra}}
    & \multirow{2}{*}{2023.06}
    & LLaVA-Instruct-150K 
    & \multirow{2}{*}{GPT-4, Prompt, Sampling ratio: 0.5} & \#image-text pairs: 158K \\
    && Flickr30K && \#images: 31.7K \hspace{0.5em} \#captions: 5 per image\\
    && Shikra-RD* && \#qustion-answer pairs: 5.9K \\
    \midrule
    
    \multirow{1}{*}{DLP\cite{jian2024bootstrapping}}
    & \multirow{1}{*}{2023.07}
    & COCO 
    & \multirow{1}{*}{-} & \#images: 330K \hspace{0.5em} \#captions: 5 per image \\
    \midrule

    \multirow{1}{*}{ChatSpot\cite{zhao2023chatspot}}
    & \multirow{1}{*}{2023.07}
    & Visual Genome 
    & \multirow{1}{*}{Unification, GPT-4, Prompt} & \#images: 108K \hspace{0.5em} \#region descriptions: 5.4M \hspace{0.5em} \#image-question pairs: 1.7M \\
    \midrule

    \multirow{2}{*}{MiniGPT-5\cite{zheng2023minigpt5}}
    & \multirow{2}{*}{2023.10}
    & VIST 
    & \multirow{2}{3cm}{\centering Flexible framework for various task, Placeholders for images -> Text prompts } & \#images: 210K \hspace{0.5em} \#stories: 50K \\
    && MMDialog && \#images: 1.53M \hspace{0.5em} \#texts: 1.08M \hspace{0.5em} \#turns: 4.92M\\
    \midrule

    \multirow{6}{*}{LLaVA-1.5\cite{liu2023improved}}
    & \multirow{6}{*}{2023.10}
    & RefCOCO 
    & \multirow{6}{3cm}{\centering Merging and concatenating, Splitting, Sampling} & \#images: 20K \hspace{0.5em} \#captions: 142K \\
    && GQA && \#images: 113K \hspace{0.5em} \#questions: 22M \\
    && OCR-VQA && \#images: 207K \hspace{0.5em} \#image-question pairs: 1M \\
    && TextCaps && \#images: 28.4K \hspace{0.5em} \#captions: 5 per image \\
    && Visual Genome && \#images: 108K \hspace{0.5em} \#region descriptions: 5.4M \hspace{0.5em} \#image-question pairs: 1.7M \\
    &&\text{llava\_v1\_5\_mix665k}* && \#image-text pairs: 665K \\
    \midrule

    \multirow{2}{*}{MiniGPT-v2\cite{chen2023minigptv2}}
    & \multirow{2}{*}{2023.10}
    & LLaVA-Instruct-150K 
    & \multirow{2}{2.7cm}{\centering Object parsing and grounding, Selecting captions} & \#image-text pairs: 158K \\
    && Flickr30K && \#images: 31.7K \hspace{0.5em} \#captions: 5 per image\\
    \midrule

    \multirow{3}{*}{CogVLM\cite{wang2024cogvlm}}
    & \multirow{3}{*}{2023.11}
    & LLaVA-Instruct-150K 
    & \multirow{3}{2.5cm}{\centering Translation into Chinese, Correcting and retranslation} & \#image-text pairs: 158K \\
    && \text{cc\_sbu\_align} && \#image-text pairs: 3.5K \\
    && CogVLM-SFT-311K* && \#images: 155K \hspace{0.5em} \#image-text pairs: 311K \\
    \midrule

    \multirow{3}{*}{DRESS\cite{chen2023dress}}
    & \multirow{3}{*}{2023.11}
    & LLaVA-Instruct-150K 
    & \multirow{3}{3.4cm}{\centering Partition the multi-turn into separate turns, LLM filtering, LLM-Human-in-the-Loop process} & \#image-text pairs: 158K \\
    && COCO && \#images: 330K \hspace{0.5em} \#captions: 5 per image \\
    && VLSafe* && \#image-text pairs: 5.8K \\
    \midrule

    \multirow{1}{*}{VILA\cite{lin2023vila}}
    & \multirow{1}{*}{2023.12}
    &\text{llava\_v1\_5\_mix665k} 
    & \multirow{1}{*}{-} & \#image-text pairs: 665K \\
    \midrule

    \multirow{9}{*}{ShareGPT4V\cite{chen2023sharegpt4v}}
    & \multirow{9}{*}{2023.11}
    & COCO 
    & \multirow{9}{*}{GPT-4, Data-specific prompt, Data replacement} & \#images: 330K \hspace{0.5em} \#captions: 5 per image \\
    && LAION2B-en && \#image-text pairs: 2.32B \\
    && CC3M && \#image-text pairs: 3.3M \\
    && SBU Captions && \#image-text pairs: 1M \\
    && SAM && \#images: 11M \hspace{0.5em} \#segmentation masks: 1.1B \\
    && TextCaps && \#images: 28.4K \hspace{0.5em} \#captions: 5 per image \\
    && WikiArt && \#images: 81.4K \hspace{0.5em} \#artist classes: 129 \hspace{0.5em} \#genre classes: 11 \hspace{0.5em} \#style classes: 27 \\
    &&\text{llava\_v1\_5\_mix665k} && \#image-text pairs: 665K \\
    && ShareGPT4V-1.2M* && \#image-text pairs: 1.2M \\
    \midrule

    \multirow{6}{*}{GLaMM\cite{rasheed2023glamm}}
    & \multirow{6}{*}{2023.11}
    & RefCOCO 
    & \multirow{6}{2.6cm}{\centering Re-purposing of data sources, Manually annotation} & \#images: 20K \hspace{0.5em} \#captions: 142K \\
    && RefCOCO+ && \#images: 20K \hspace{0.5em} \#captions: 141K \\
    && RefCOCOg && \#images: 25.8K \hspace{0.5em} \#captions: 95K \\
    && Visual Genome && \#images: 108K \hspace{0.5em} \#region descriptions: 5.4M \hspace{0.5em} \#image-question pairs: 1.7M \\
    && LLaVA-Instruct-150K && \#image-text pairs: 158K \\
    && GranD-f* && \#image-text pairs: 214K \\
    \midrule

    \multirow{6}{*}{PVIT\cite{chen2023position}}
    & \multirow{6}{*}{2023.08}
    & GQA 
    & \multirow{6}{*}{ChatGPT, Prompt, Multi-turn data} & \#images: 113K \hspace{0.5em} \#questions: 22M \\
    && VCR && \#images: 110K \hspace{0.5em} \#questions: 290K \\
    && COCO && \#images: 330K \hspace{0.5em} \#captions: 5 per image \\
    && Visual Genome && \#images: 108K \hspace{0.5em} \#region descriptions: 5.4M \hspace{0.5em} \#image-question pairs: 1.7M \\
    && COCO-Text && \#images:63.6K \hspace{0.5em} \#labeled text regions: 173K\\
    && PVIT* && \#image-text pairs: 13.7M (Stage 1) \hspace{0.5em} \#image-question pairs: 10.4K (Stage 2) \\
    \midrule

    \multirow{9}{*}{TextMonkey\cite{liu2024textmonkey}}
    & \multirow{9}{*}{2024.03}
    & COCO-Text 
    & \multirow{9}{3.2cm}{\centering Structured data (documents, tables, charts), 5\% of pre-train data, } & \#images:63.6K \hspace{0.5em} \#labeled text regions: 173K\\
    && TextOCR && \#images:28.1K \hspace{0.5em} \#texts: 903K \\
    && HierText && \#images:11.6K \hspace{0.5em} \#texts: 1.2M \\
    && TextVQA && \#images: 28.4K \hspace{0.5em} \#questions: 45.3K \\
    && MLT && \#image-text pairs: 10K \\
    && ChartQA && \#image-text pairs: 20.8K \hspace{0.5em} \#tables: 20.8K \\
    && DocVQA && \#images: 12K \hspace{0.5em} \#questions: 50K\\
    && InfoVQA && \#images: 5K \hspace{0.5em} \#questions: 30K\\
    &&\text{Monkey\_Data}* && \#image-text pairs: 409.1K \\   
    \bottomrule
    \end{tabular}
    \label{table:sft:image}
\end{table}

\begin{table}
    
    \centering
    \caption{A detailed list of datasets and processing methods for different models with video modality during fine-tuning stage. * indicates the dataset is newly generated using certain method within a respective model, while the other datasets (without *) serve as the original data sources for that model. - denotes directly using original data sources without additional processing. \#Examples denotes the statistics for each dataset.}
    \tiny
    \setstretch{0.95}
    \begin{tabular}{cclcl}
    \toprule
    
    \textbf{Models} & \textbf{Time} & \multicolumn{1}{c}{\textbf{Finetuning Datasets}} & \multicolumn{1}{c}{\textbf{Processing Method}} & \multicolumn{1}{c}{\textbf{\#Examples}} \\
    \midrule
    
    \multirow{2}{*}{VideoChat\cite{li2023videochat}}
    & \multirow{2}{*}{2023.05}
    & WebVid 
    & \multirow{2}{2.2cm}{\centering GPT-4, Prompt, Randomly choosing} & \#video-text pairs: 10M \\
    && VideoChat* && \#video-description pairs: 7K \hspace{0.5em} \#video-conversation pairs: 4K \\
    \midrule
    
    \multirow{1}{*}{Video-LLaMA\cite{lin2023video}}
    & \multirow{1}{*}{2023.01}
    & VideoChat 
    & \multirow{1}{*}{-} & \#video-description pairs: 7K \hspace{0.5em} \#video-conversation pairs: 4K \\
    \midrule
    
    \multirow{2}{*}{Video-ChatGPT\cite{maaz2023video}}
    & \multirow{2}{*}{2023.01}
    & ActivityNet 
    & \multirow{2}{2.75cm}{\centering Human-assisted annotations, GPT, Semi-automatic annotation} 
    & \#videos: 20K \hspace{0.5em} \#texts: 100K \\
    && VideoInstruct100K* && \#video-instruction pairs: 100K \\      
    \bottomrule
    \end{tabular}
    \label{table:sft:video}
\end{table}

\begin{table}
    \centering
    \caption{A detailed list of datasets and processing methods for different models with audio modality during the fine-tuning stage. * indicates the dataset is newly generated using certain methods within a respective model, while the other datasets (without *) serve as the original data sources for that model. The datasets in other modals used for generating multi-modal datasets (including audio) are also shown in this table. - denotes directly using original data sources without additional processing. \#Examples denote the statistics for each dataset.}
    \tiny
    \setstretch{0.95}
    \begin{tabular}{cclcl}
    \toprule
    
    \textbf{Models} & \textbf{Time} & \multicolumn{1}{c}{\textbf{Finetuning Datasets}} & \multicolumn{1}{c}{\textbf{Processing Method}}& \multicolumn{1}{c}{\textbf{\#Examples}} \\
    \midrule
    
    \multirow{4}{*}{BuboGPT\cite{zhao2023bubogpt}}
    & \multirow{4}{*}{2023.07}
    & Clotho 
    & \multirow{4}{*}{GPT-4, Prompt}& \#audios: 5K \hspace{0.5em} \#captions: 24K \\
    && VGGSS && \#video-audio pairs: 5K \\
    && Clotho-Detail* && \#audio-caption pairs: 3.9K \hspace{0.5em} \#tokens: 207K \\
    && VGGSS-Instruction-Tuning* && \#audio-image-caption pairs: 5.1K \\
    \midrule
    
    \multirow{4}{*}{X-LLM\cite{chen2023x}}
    & \multirow{4}{*}{2023.05}
    & AISHELL-2 
    & \multirow{4}{2cm}{\centering Manually selecting, ChatGPT}& \#reading-speech audios: 1000 hours \\
    && VSDial-CN && \#automatic speech recognition (ASR) samples: 1.2M \\
    && \text{cc\_sbu\_align (Image-Text)} && \#image-text pairs: 3.5K \\
    && ActivityNet (Video-Text)  && \#videos: 20K \hspace{0.5em} \#texts: 100K \\
    \midrule

    \multirow{5}{*}{NExT-GPT\cite{wu2023next}}
    & \multirow{5}{*}{2023.09}
    & WebVid (Video-Text) 
    & \multirow{5}{2cm}{\centering External resources, GPT-4, Prompt}& \#video-text pairs: 10M \\
    && AudioCaps && \#audio-text pairs: 46K \\
    && CC3M (Image-Text) && \#image-caption pairs:  3.3M \\
    && T2M* && \#images: 4.9K \hspace{0.5em} \#videos: 4.9K \hspace{0.5em} \#audios: 4.9K \hspace{0.5em} \#instances: 14.7K \\
    && MosIT* && \#images: 4K \hspace{0.5em} \#videos: 4K \hspace{0.5em} \#audios: 4K \hspace{0.5em} \#instances: 5K \\
    \midrule
    
    \multirow{3}{*}{X-InstructBLIP\cite{panagopoulou2023x}}
    & \multirow{3}{*}{2023.11}
    & AudioCaps 
    & \multirow{3}{2cm}{\centering Automatic generation, Prompt} & \#audio-text pairs: 46K \\
    && AudioCapsQA* && \#audio QA pairs: 25.4K \hspace{0.5em} \#unique questoins: 10.8K \\
    && DisCRn* && \#audio-video pairs: 8.8K \\      
    \midrule

    \multirow{2}{*}{Qwen-Audio\cite{bai2023qwen}}
    & \multirow{2}{*}{2023.11}
    & \multirow{2}{*}{Self-constructed}
    & \multirow{2}{2cm}{\centering Manual annotation, GPT-3.5, Data Mixing} & \multirow{2}{*}{\#audio-text pairs: 20K} \\ 
    \\
    \bottomrule
    \end{tabular}
    \label{table:sft:audio}
\end{table}